\newif\ifeccv \eccvtrue
\newif\ifdraft \draftfalse
\newif\ifanonymous \anonymousfalse
\newif\ifarxiv \arxivtrue

\documentclass[runningheads]{ECCV22/llncs}
\usepackage{graphicx} 

\usepackage{tikz}
\usepackage{comment}
\usepackage{amsmath,amssymb} 
\usepackage{color}

\usepackage[accsupp]{axessibility}  

\ifanonymous
\usepackage{ECCV22/ruler}
\usepackage[width=122mm,left=12mm,paperwidth=146mm,height=193mm,top=12mm,paperheight=217mm]{geometry}
\fi

\usepackage{comment}
\usepackage{caption}
\usepackage{stackengine} 

\ifundef{\ifanonymous}
{\newif\ifanonymous \anonymousfalse}
{}

\ifundef{\ifdraft}
{\newif\ifdraft \draftfalse}
{}

\ifundef{\ifeccv}
{\newif\ifeccv \eccvfalse}
{}

\ifundef{\ifcvpr}
{\newif\ifcvpr \cvprfalse}
{}

\ifundef{\ifcvprreview}
{\newif\ifcvprreview \cvprreviewfalse}
{}

\ifundef{\ifappendix}
{\newif\ifappendix \appendixfalse}
{}

\ifundef{\ifarxiv}
{\newif\ifarxiv \arxivfalse}
{}

\ifundef{\ificcv}
{\newif\ificcv \iccvfalse}
{}

\ifundef{\iftog}
{\newif\iftog \togfalse}
{}

\ifundef{\ificcvfinal}
{\newif\ificcvfinal \iccvfinalfalse}
{}

\ifarxiv
\appendixtrue
\fi

\iftog
\appendixtrue
\fi

\ifanonymous
{\global\iccvfinalfalse}
\else
{\global\iccvfinaltrue}
\fi

\ifcvpr{
\newcommand{\citet}{\cite}
\newcommand{\keywords}{\null}
}\fi

\ifeccv
\else
\usepackage{times}
\fi
\usepackage{ifthen}
\usepackage{cancel}
\usepackage{epsfig}
\usepackage{graphicx}
\usepackage{amsmath}
\iftog {} \else \usepackage{amssymb} \fi 

\usepackage{xcolor}
\usepackage{enumitem}
\usepackage{wrapfig}
\usepackage{float}
\usepackage{xspace} 
\usepackage{orcidlink}

\newcommand{\bK}{{\bf K}}
\newcommand{\bQ}{{\bf Q}}

\newcommand{\bV}{{\bf V}}
\newcommand{\bJ}{{\bf J}}
\newcommand{\bP}{{\bf P}}
\newcommand{\bL}{{\bf L}}
\newcommand{\bT}{{\bf T}}

\newcommand{\bff}{{\bf f}}

\newcommand{\bq}{{\bf q}}
\newcommand{\br}{{\bf r}}
\newcommand{\bss}{{\bf s}}

\newcommand{\Loss}{\mathcal{L}}

\newcommand{\nn}{\mathcal{N}}
\newcommand{\pp}{\mathcal{P}}
\newcommand{\qq}{\mathcal{Q}}
\newcommand{\bbe}{\mathbb{E}}
\newcommand{\bbr}{\mathbb{R}}

\iftog 
\else
\newcommand{\shortcite}{\cite}
\fi

\usepackage{url}
\usepackage{graphics}
\usepackage[normalem]{ulem} 
\usepackage{multirow}
\usepackage{tabu,stackengine}
\usepackage{wrapfig}
\usepackage{booktabs}
\usepackage{soul}
\usepackage{makecell}
\ifcvpr
\else
\usepackage{cleveref}
\crefname{section}{Sec.}{Secs.}
\Crefname{section}{Section}{Sections}
\Crefname{table}{Table}{Tables}
\crefname{table}{Tab.}{Tabs.}
\fi

\newcommand{\bluebold}[1]{{\textbf{\color{blue} #1}}}

\definecolor{applegreen}{rgb}{0.55, 0.71, 0.0}
\definecolor{burgundy}{rgb}{0.5, 0.0, 0.13}
\definecolor{calpolypomonagreen}{rgb}{0.12, 0.3, 0.17}

\ifdraft
\newcommand{\setcolor}[1]{\color{#1}}
\newcommand{\sr}[1]{{\color{violet} #1}}

\newcommand{\bg}[1]{{\color{orange} #1}}
\else
\newcommand{\setcolor}[1]{}
\newcommand{\sr}[1]{{#1}}

\newcommand{\bg}[1]{#1}
\fi

\newcommand{\algoname}{FLEX}
\newcommand{\projectpage}{\url{https://briang13.github.io/FLEX}}

\makeatletter
\iftog
\newcommand{\sectiontinyvert}{\section}
\newcommand{\paragraphtinyvert}{\paragraph}
\newcommand{\paragraphnovert}{\paragraph}
\newcommand{\subparagraphnovert}{\subparagraph}
\else

\newenvironment{myequation*}{%
\list{}{\topsep by-7pt
\mathdisplay@push
  \st@rredtrue \global\@eqnswfalse
  \mathdisplay{equation}%
}}{%
  \endmathdisplay{equation}%
  \mathdisplay@pop
  \ignorespacesafterend
  \endlist
}

\newenvironment{myabstract}{%
      \list{}{\advance\topsep by-7pt\relax\small
      \leftmargin=0.95cm
      \labelwidth=\z@
      \listparindent=\z@
      \itemindent\listparindent
      \rightmargin\leftmargin}\item[\hskip\labelsep
                                    \bfseries\abstractname]}
    {\endlist}

\newcommand{\sectiontinyvert}{
  \@startsection{section}               
  {1}                                   
  {\z@}                                 
  {10pt \@plus 3pt \@minus 0pt}         
  {5pt \@plus 2pt \@minus 0pt}          
  {\normalfont\large\bfseries\boldmath\rightskip=\z@ \@plus 8em\pretolerance=10000}          
}

\newcommand{\subsectiontinyvert}{
  \@startsection{subsection}            
  {2}                                   
  {\z@}                                 
  {10pt \@plus 3pt \@minus 0pt}         
  {5pt \@plus 2pt \@minus 0pt}          
  {\normalfont\normalsize\bfseries\boldmath\rightskip=\z@ \@plus 8em\pretolerance=10000}     
}
\newcommand\subsubsectiontinyvert{
    \@startsection{subsubsection}
    {3}
    {\z@}
    {10pt \@plus 3pt \@minus 0pt}%
    {-0.5em \@plus -0.22em \@minus -0.1em}%
    {\normalfont\normalsize\bfseries\boldmath}}

\newcommand{\paragraphtinyvert}{%
  \@startsection{paragraph}{4}%
  {\z@}{1ex \@plus 0.0ex \@minus 0.2ex}{-1em}
  {\normalfont\normalsize\bfseries\boldmath} 
}
\newcommand{\paragraphnovert}{%
  \@startsection{paragraph}{4}%
  {\z@}{0ex \@plus 0ex \@minus 0ex}{-0.5 em}%
  {\normalfont\normalsize\bfseries}%
}
\newcommand{\subparagraphnovert}{%
  \@startsection{subparagraph}{5}%
  {3ex}{0ex \@plus 0ex \@minus 0ex}{-1em}%
  {\normalfont\normalsize\bfseries}%
}

\newcommand{\orcid}[1]{\orcidlink{#1}}

\fi

\ificcv
\else
\DeclareRobustCommand\onedot{\futurelet\@let@token\@onedot}
\def\@onedot{\ifx\@let@token.\else.\null\fi\xspace}

\def\eg{\emph{e.g}\onedot}

\def\etal{\emph{et al}\onedot}
\fi

\makeatother

\begin{document}
\pagestyle{headings}
\mainmatter
\def\ECCVSubNumber{6730}  

\title{\algoname: Extrinsic Parameters-free Multi-view 3D Human Motion Reconstruction} 

\ifarxiv
\else
\titlerunning{ECCV-22 submission ID \ECCVSubNumber} 
\authorrunning{ECCV-22 submission ID \ECCVSubNumber} 
\author{Anonymous ECCV submission}
\institute{Paper ID \ECCVSubNumber}
\fi

\titlerunning{FLEX}

\ifanonymous 
\else
\author{
Brian Gordon\thanks{equal contribution.} \orcid{0000-0002-3016-3690} \and 
Sigal Raab$^*$  \orcid{0000-0001-6616-257X} \and 
Guy Azov  \orcid{0000-0003-2336-8601} \and \\
Raja Giryes  \orcid{0000-0002-2830-0297} \and 
Daniel Cohen-Or \orcid{0000-0001-6777-7445}
}
%
\authorrunning{Gordon and Raab et al.}
%
\institute{Tel Aviv University \\
\email{
briangordon@mail.tau.ac.il, sigalraab@tauex.tau.ac.il, guyazov@mail.tau.ac.il,
\{raja,dcor\}@tauex.tau.ac.il}
}
\fi

\maketitle
\begin{myabstract}
The increasing availability of video recordings made by multiple cameras has offered new means for mitigating occlusion and depth ambiguities in pose and motion reconstruction methods. Yet,  multi-view algorithms strongly depend on camera parameters, particularly on relative transformations between the cameras. 
Such a dependency becomes a hurdle once shifting to dynamic capture in uncontrolled settings. 
We introduce \algoname \, (\textbf{F}ree mu\textbf{L}ti-view r\textbf{E}constru\textbf{X}ion), an end-to-end extrinsic parameter-free multi-view model. FLEX is \emph{extrinsic parameter-free} (dubbed \emph{ep-free}) in the sense that it does not require extrinsic camera parameters. 
Our key idea is that the 3D angles between skeletal parts, as well as bone lengths, are invariant to the camera position. 
Hence, learning 3D rotations and bone lengths rather than locations allows for predicting common values for all camera views. 
Our network takes multiple video streams, learns fused deep features through a novel multi-view fusion layer,
and reconstructs a single consistent skeleton with temporally coherent joint rotations.
We demonstrate quantitative and qualitative results on three public data sets, and on multi-person synthetic video streams captured by dynamic cameras.
We compare our model to state-of-the-art methods that are not ep-free and show that in the absence of camera parameters, we outperform them by a large margin while obtaining comparable results when camera parameters are available.
Code, trained models, and other materials are available on \ifanonymous{our project page.}\else{\projectpage.}\fi

\ifeccv
\fi

\end{myabstract}

\begin{figure}[t]
\setlength{\abovecaptionskip}{5pt plus 3pt minus 2pt}
\setlength{\belowcaptionskip}{-20pt plus 3pt minus 2pt}

\centering
\begin{minipage}[b]{0.52\linewidth}
    \centering
    \includegraphics[width=\textwidth,height=3cm]{./images/football_results.pdf}
    \caption{Results on the KTH Multi-view Football II dataset~\cite{footballDS}, in occluded and blurry scenes with dynamic cameras.
    }
    \label{fig:football_teaser}
\end{minipage}
\hfill
\begin{minipage}[b]{0.44\linewidth}
    \centering
    \includegraphics[width=\textwidth,height=3cm]{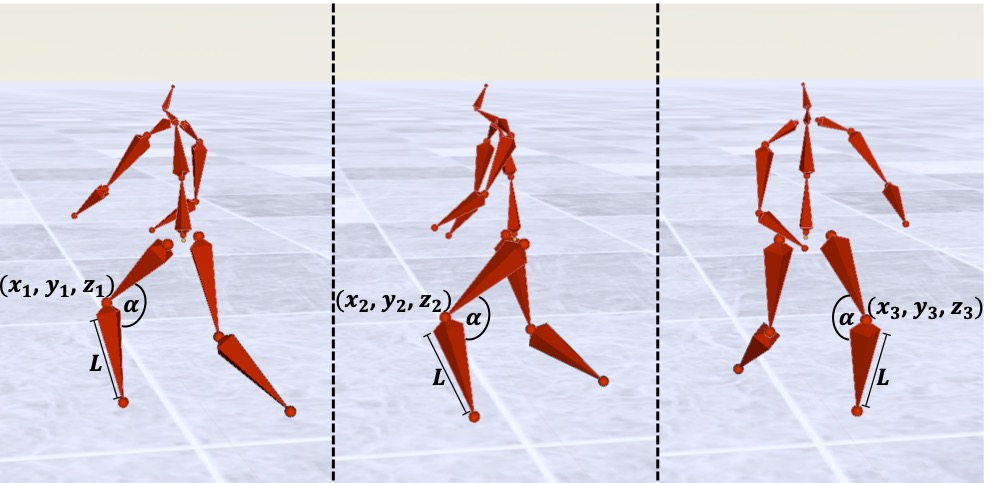}
    \caption{3D locations vary across axis systems while 3D rotation angles and bone lengths remain identical.
    }
    \label{fig:skeleton_angles}
\end{minipage}
\end{figure}

\sectiontinyvert{Introduction}
Human motion reconstruction is the task of associating a skeleton with
temporally coherent joint locations and rotations. 
Acquiring accurate human motion in a controlled setting, using motion capture systems with adequate sensors is a tedious and expensive procedure that cannot be applied for capturing spontaneous activities, such as sporting events.
Motion reconstruction from RGB cameras is low-cost and non-intrusive, but is an uncontrolled setup. Thus, while being simple, it has technical challenges that are worsened by occlusion and depth ambiguity. 
Using multiple cameras may alleviate these difficulties as different views may compensate for occlusion and be used for mutual consistency. 

Recently, there has been a significant progress in using deep learning for pose and motion reconstruction \cite{pavlakos2018ordinal,liu2019improving,sarandi2020metric,shi2020motionet,pavllo20193d,kocabas2020vibe,mehta2020xnect}. Most of these methods work in a monocular setting, but a growing number of works learn a multi-view setting \cite{iskakov2019learnable,tu2020voxelpose,qiu2019cross,he2020epipolar,dong2019fast,rhodin2018learning}. However, these approaches 
\sr{depend on} the 
relative position 
between the cameras, derived from extrinsic camera parameters, \sr{and assume they} are given.
%
In the lack of extrinsic parameters, several works estimate them~\cite{chu_and_pan_semisupervised,kocabas2019selfsupervised}, 
but at the cost of innate inaccuracy of estimated values.
While camera parameters are often given in multi-view datasets, they are rarely given in dynamic capture environments. We refer to cameras as \emph{dynamic} if they occasionally move during video capture, such that their extrinsic parameters and their inter-camera relative positions are not fixed. \bg{An example of such a camera is the SkyCam~\cite{skycam}, commonly used in sports events.}



\begin{wrapfigure}{r}{0.3\textwidth}
\setlength{\abovecaptionskip}{-48pt plus 3pt minus 2pt}
\setlength{\belowcaptionskip}{0pt plus 3pt minus 2pt}
\caption*{}

  \centering
    \includegraphics[width=0.3\textwidth]{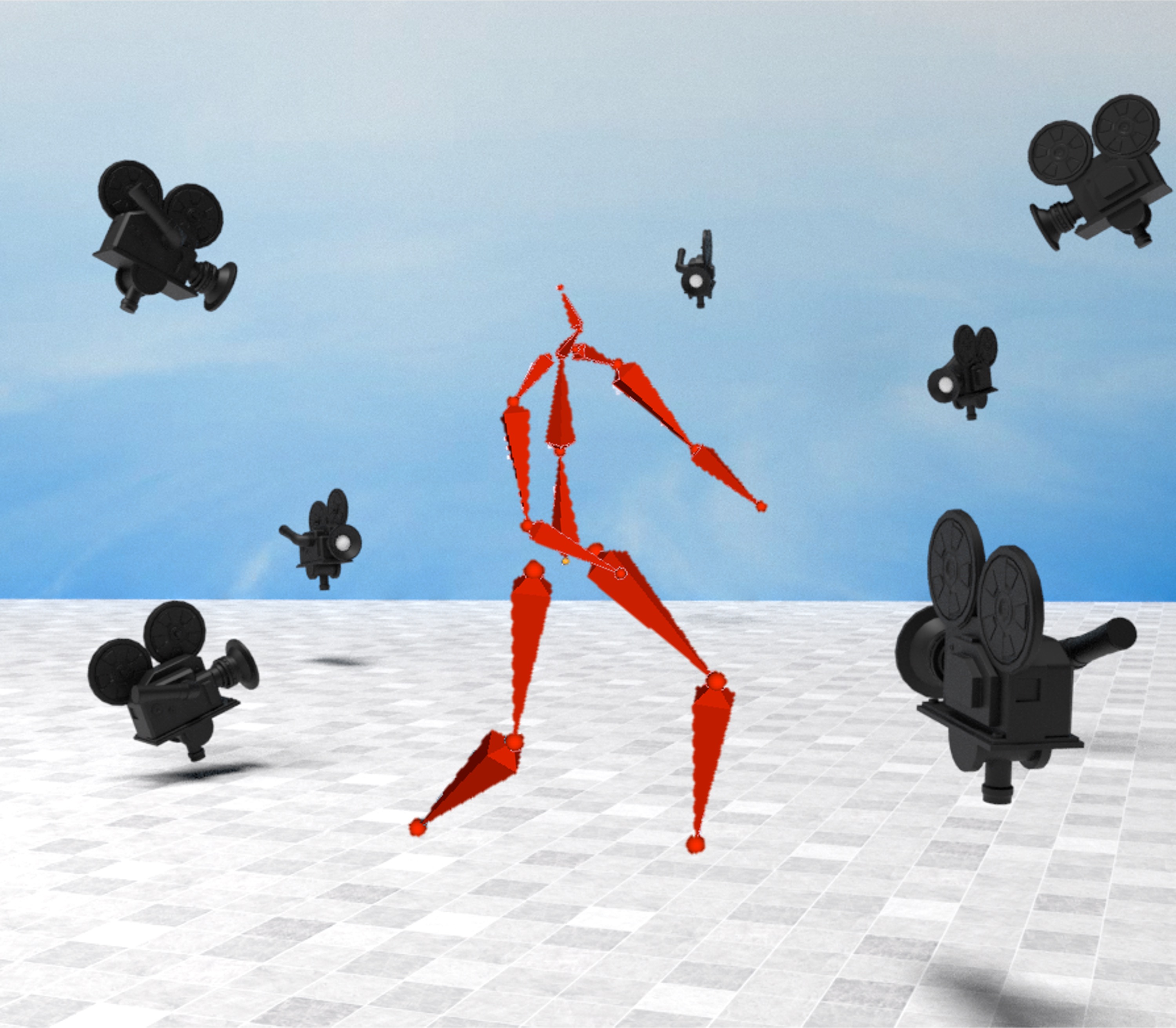}
   \label{fig:skeleton_cameras}
   
\setlength{\abovecaptionskip}{-45pt plus 3pt minus 2pt}
\setlength{\belowcaptionskip}{-5pt plus 3pt minus 2pt}
\caption*{}
\end{wrapfigure}

This work introduces an extrinsic parameter-free (dubbed \emph{ep-free}) multi-view motion reconstruction method, whose setting is illustrated in the inset to the right. 
%
Our method builds upon a new conceptual observation that uses the well-known joint rotations and bone lengths, to free us from the burdening dependency on extrinsic camera parameters.
Our approach relies on a key insight that joint rotations and bone lengths are identical for all views. 
That is, the 3D angle between skeletal parts is invariant to the camera position.
%
We train a neural network to predict 3D joint angles and bone lengths \emph{without} using the extrinsic camera parameters, 
neither in training nor in test time.
Predicting motion rather than locations is not a novel idea by itself. The innovation of our work is in the way we use motion to bypass the need for camera parameters.
The input from multiple cameras is integrated by a novel fusion layer that implicitly promotes joints detected by some cameras and demotes joints detected by others, 
hence  mitigating occlusion and depth ambiguities.

Our model, named FLEX, is an end-to-end deep convolutional network. Its input is multi-view 2D joints that are either given or extracted using a 2D pose estimation technique. 
FLEX employs multi-view blocks with cross-view attention on top of a monocular baseline~\cite{shi2020motionet}, 
and 
uses temporal information over a video of arbitrary length, thus obtaining temporal consistency.

We evaluate FLEX qualitatively and quantitatively using the Human3.6M~\cite{h36m_pami,IonescuSminchisescu11}, the KTH Multi-view Football II~\cite{footballDS} and the Ski-Pose PTZ-Camera~\cite{ski_ptz} datasets. 
\Cref{fig:football_teaser} demonstrates qualitative results, and more are depicted in \Cref{sec:experiments} and in the \ifarxiv{appendix}\else{supplementary material}\fi.
%
FLEX is also applied on synthetic videos. We have generated these videos using Mixamo~\shortcite{mixamo} and Blender~\shortcite{blender},  to mitigate the lack of a multi-person video dataset that is captured by dynamic cameras, and created them such that they contain severe inter-person occlusions.

We compare performance with state-of-the-art methods that are not ep-free and show comparable results. To simulate an ep-free setting, we perturb ground-truth camera parameters or use works that estimate them. We show that in an ep-free setting, our model outperforms state-of-the-art by a large margin. 

Our main contributions are twofold: (i) a network that reconstructs motion and pose in a multi-view setting with unknown extrinsic camera parameters, and (ii) a novel fusion layer with a multi-view convolutional layer combined with a multi-head attention mechanism over a number of views.

\sectiontinyvert{Related work}

\label{sec:related}

\paragraphtinyvert{Pose Estimation using a Single View}
Pose estimation receives significant interest in computer vision. Before the deep era, it was approached using heuristics such as physical priors \cite{Sarafianos_2016}. 
The emergence of deep learning and large datasets~\cite{h36m_pami,CMU:mocap,3DPWvonMarcard2018,footballDS}, have led to significant advances. 
Pose estimation methods can generally be divided into two groups. The first  infers 3D locations directly from images or videos \cite{firstSingleViewCNN,Pavlakos_coarse_to_fine,zhu2020nbaplayers,tekin2016structured,sun2018integral,cheng20203d,habibie2019wild}. 
The second, aka \emph{lifting}, applies two steps: (i) estimating 2D poses and (ii) lifting them to 3D space~\cite{martinez2017simple,pavllo20193d,fang2018learning,sym12071116,liu2019improving,Hossain_2018,Shan2021ImprovingRA}.  
The first group benefits from directly using images, which are more descriptive compared to 2D joint locations. The second gains from using intermediate supervision.
%
%
Recently, transformers and convolutional graph based methods were shown to improve performance~\cite{llopart2020liftformer,lin2020endtoend,liu2020attention,li2021exploiting,wang2020motion,hu2021conditional,Human_Mesh_Recovery_from_Multiple_Shots}.

\paragraphtinyvert{Pose Estimation using Multiple Views} 
The growing availability of synchronized video streams taken by multiple cameras has contributed to the emergence of multi-view algorithms. Such algorithms exploit the diversity in camera views to predict more accurate 3D poses. All works described below predict pose and many of them analyze each frame individually. On the other hand, our model, FLEX, reconstructs motion and exploits temporal information. 

Most works in the multi-view setting rely on lifting from 2D to 3D space.
Early works~\cite{Belagiannis_2016_IEEE,detection_complete_graphs,psm_2013} estimate the input 2D pose from single images, while later works~\cite{dong2019fast,qiu2019cross,iskakov2019learnable,he2020epipolar,kocabas2019selfsupervised,chen2019weaklysupervised,chu_and_pan_semisupervised,rhodin2018learning,kadkhodamohammadi2019generalizable,chu2021partaware} obtain the 2D pose by running a CNN over 2D poses given in multiple views; resulting in an increase in 2D pose prediction accuracy.
After estimating the 2D poses, most works apply heuristics such as triangulation or pictorial structure model (PSM). FLEX is one of the few works \cite{iskakov2019learnable,tu2020voxelpose} that present an end-to-end model. 

Several methods 
use multi-view data to improve the 2D pose estimation.
Some use the camera parameters to find the matching epipolar lines 
such that features gathered from several cameras are aggregated \cite{qiu2019cross,he2020epipolar}. 
Chen \etal~\shortcite{chen2019weaklysupervised} learn a geometric representation in latent space with an encoder-decoder.

Several works~\cite{kocabas2019selfsupervised,chu_and_pan_semisupervised,wandt2020canonpose,tome2018rethinking} use self-supervision, hence need no 3D ground-truth.
Their main idea is to project the predicted 3D joints (using real or estimated camera parameters) and expect consistency with 2D input joints.
%
Recent techniques \cite{huang2019deepfuse,zhang2020fusing} exploit more sensors, such as IMU, during data capturing.
 

Current state-of-the-art results are attained by Iskakov \etal~\cite{iskakov2019learnable}, Tu \etal~\cite{tu2020voxelpose} and Reddy \etal~\cite{Reddy2021TesseTrackEL}. They use end-to-end networks, and present a volumetric approach, where 2D features are un-projected from individual views to a common 3D space, using camera parameters.
Sun \etal~\shortcite{sym12071116} show that synthetic generation of additional views helps produce more accurate lifting. 

At inference time, some of the aforementioned works expect monocular inputs \cite{sym12071116,he2020epipolar,chen2019weaklysupervised,chu_and_pan_semisupervised} and some, including FLEX, get multi-view inputs \cite{iskakov2019learnable,tu2020voxelpose,qiu2019cross}. The advantage of the first is the use of monocular data that is more common, and of the second is better results on multi-view settings. 

Epipolar Transformers~\cite{he2020epipolar} attend to spatial locations on an epipolar line in a \emph{single} view and 
query it using one joint in a query view. 
A concurrent work, TransFusion~\cite{ma2021transfusion},
\sr{applies a transformer on inter and intra-view features.}

In the absence of camera parameters,
most of the methods cannot be used. Some estimate rotation assuming the translation is given \cite{kocabas2019selfsupervised,bachmann2019motion} or engage an extra effort to estimate the camera parameters \cite{chu_and_pan_semisupervised,wandt2020canonpose,chen2021deductive,usman2021metapose,human_pose_calib_unsync}. 
Such an effort is not required by FLEX as it uses no camera parameters whatsoever. 

\paragraphtinyvert{Rotation and Motion Reconstruction}
Pose estimation may suffice for many applications; however, pose alone does not fully describe the motion and the rotations associated with the joints. \emph{Rotation reconstruction} relates to the prediction of joint rotation angles, while \emph{motion reconstruction} requires the prediction of bone lengths associated with them.
Many works explore the task of \emph{3D shape recovery}~\cite{SMPL:2015,Kanazawa:2018,Kolotouros:2019:ICCV,kocabas2020vibe,kissos2020weak,yoshiyasu2018skeleton,kanazawa2019learning,habermann2020deepcap,luo20203d,choi2021static}, focusing  on human mesh prediction along with joint rotations. Most of them do not guarantee temporal coherence, \eg, bone length may vary across time frames. 

Other works~\cite{pavllo2018quaternet,mao2020learning} 
focus on motion generation. Given a series of human motions, they predict future motions, using various techniques such as temporal supervision and graph convolutional networks (GCN).
Similar to us, human motion reconstruction methods~\cite{zhou2016deep,mehta2020xnect,shi2020motionet,video_motion_capture_cpm} focus on the temporal coherence of the body, where the bone lengths are fixed over time and rotations are smooth.

\sectiontinyvert{Extrinsic Parameter-free multi-view model}
The premise of our work is that 3D joint rotations and bone lengths are view-independent values. 
For example, the 3D angle between, say, the thigh and the shin, as well as the length of these bones, are fixed, no matter which camera transformation is used. On the other hand, joint locations  differ  for each camera transformation, as seen in \Cref{fig:skeleton_angles}. 
Our key idea is to directly predict joint 3D angles and bone lengths without using the extrinsic camera parameters, during both training and test time. 
\textit{Extrinsic} parameters correspond to the rotation and translation (aka transformation) from 3D real world axes into 3D camera axes.
A formal definition of the camera parameters can be found in 
\ifappendix{Appendix \ref{sec:cam_param_technical}.}
\else{the sup. mat.}
\fi

Our method takes multi-view sequences of 2D poses and estimates the motion of the observed human.
The 2D poses are either given or extracted using a prediction technique. Having multi-view data compensates for the inherent inaccuracy of 2D pose estimation algorithms.
Many methods estimate view-dependent 2D joint positions and then lift them to 3D by transforming them into a shared space. Such transformations require acquaintance of the relative position (rotation and translation) between the cameras, which is derived from the extrinsic camera parameters. 
Our model directly predicts 3D rotations and bone lengths, which are agnostic to camera transformation. The predicted values are shared by all views, so there is no need for extrinsic parameters information.

Pose estimation methods may mitigate the lack of extrinsic  parameters by estimating them~\cite{chu_and_pan_semisupervised,kocabas2019selfsupervised}.
Yet, this has two drawbacks: 
(i) most approaches perform the estimation in a prepossessing step that breaks the end-to-end computation, and 
(ii) the estimated parameters are never exact and typically lead to a performance drop, as we show in \Cref{sec:experiments}.

\begin{figure*}[thb]
\centering
\includegraphics[width=0.9\textwidth]{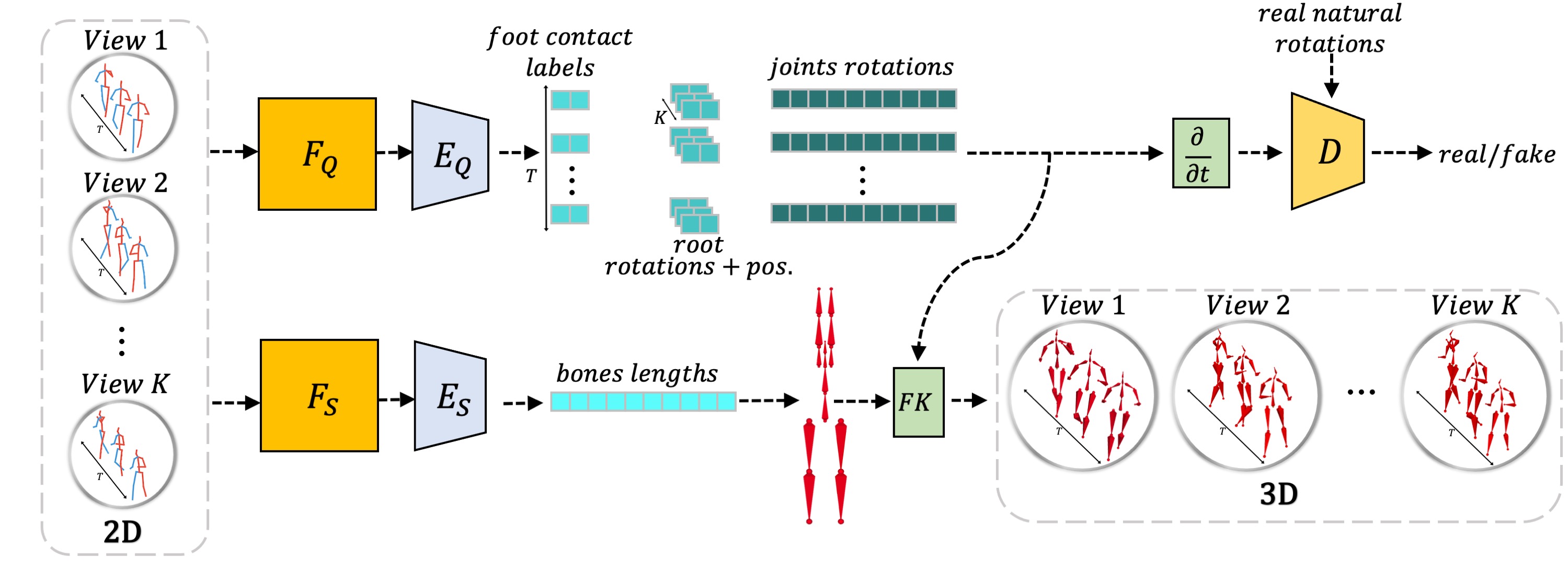}
\setlength{\abovecaptionskip}{0pt plus 3pt minus 2pt}
\setlength{\belowcaptionskip}{-15pt plus 3pt minus 2pt}
\caption{
%
%
%
%
FLEX takes multi-view temporal sequences of 2D poses and their confidence values. 
It uses two encoders, $E_Q$ and $E_S$, to extracts per-frame 3D rotations and foot contact labels, per-view and per-frame 3D root transformations, and one static skeleton. 
A discriminator $D$ monitors the temporal differences of rotation angles,
and a forward kinematic layer, $FK$, 
combines encoders' outputs into 3D joint locations.
%
\sr{These outputs depict \emph{one} human, transformed into the axis systems of $K$ cameras, to be compared with $K$ sets of  ground-truth values.}
}

\label{fig:architecture_concept}
\end{figure*}

Our architecture leverages Shi \etal~\cite{shi2020motionet} and is illustrated in high-level terms in \Cref{fig:architecture_concept}. FLEX is an end-to-end network that maps 2D joint positions, extracted from multiple synchronized input videos, into two separate components: (i) a sequence of 3D joint rotations, global root positions and foot contact labels (upper branch in the figure); this sequence is skeleton-independent and varies per frame; and (ii) a single, symmetric, 3D skeleton, represented by its bone lengths (lower branch in the figure). We can combine these two components into a complete description of a motion and use it for 3D animation tasks without further processing or inverse kinematics (IK).

In addition to being free of extrinsic parameters, our model does not use intrinsic parameters at all, at the cost of an up-to-scale global skeleton position. While FLEX removes the need for extrinsics, it uses the common weak perspective assumption~\cite{kissos2020weak} for intrinsics; in particular for mitigating the lack of focal length. Indeed, some works seek to mitigate the lack of intrinsic parameters~\cite{shimada2021neural,habermann2020deepcap,kissos2020weak} whereas this is not the focus of our work.
 In \Cref{sec:experiments} we show that using a customary weak perspective we attain an accurate global position.


The terms \emph{motion}, \emph{pose}, \emph{reconstruction} and \emph{estimation} are used in various contexts in the literature. To avoid confusion, we define \emph{motion} as one set of bone lengths associated with temporally coherent 3D joint rotations, and \emph{pose} as a temporal sequence of 3D joint locations.
We use the term \emph{reconstruction} rather than \emph{estimation}, as the latter often describes 2D spatial motion. \sr{A weakly related term, \emph{pose tracking}, associates poses to identities in a multi-person setting.}

Motion data, and in particular rotations rather than positions, are required in animation platforms and game engines.
FLEX directly outputs a kinematic skeleton, which is a complete, commonly used, motion representation. 
On the other hand, methods that predict joint positions, rely on IK to associate a skeleton with joint rotations. 
IK is slow, non-unique, and prone to temporal inconsistencies and unnatural postures.
Moreover, methods that only predict pose cannot guarantee the consistency of bone lengths across frames.


\subsectiontinyvert{Architecture} \label{sec:architecture}

We start with a high-level description of the architecture (see \Cref{fig:architecture_concept}). 
%
The inputs are $K$ synchronized video streams of $T$ frames each. For each video stream, we obtain 2D joints, which are either the ground-truth of a dataset or the output of a 2D pose estimation algorithm. Our network is agnostic to the way those 2D joints were obtained. In addition, 
each estimated joint is associated with a confidence value.
The confidence value plays an important role in balancing between visible and occluded joints.

Our model takes input from all views, aggregates it, and streams it into two independent fusion layers $F_S$ and $F_Q$, followed by encoders $E_S$ and $E_Q$, respectively. The two fusion layers differ in some architectural details, but share the same concept. Both aggregate data of all views and frames and fuse it to exploit characteristics that recur in views and/or frames. Each fusion layer outputs view-agnostic features that represent the target human. 

The fusion layers consist of two innovative elements, a multi-view convolutional layer and a \emph{cross-view attention} mechanism, which encodes information from all views. Our use of attention is unique, as typically attention in other works is applied mostly over pixels~\cite{khan2021transformers} and sometimes over time~\cite{llopart2020liftformer,lin2020endtoend,liu2020attention}. 
Attention over views is a novel approach, which we find only in concurrent works for other tasks, assessing human shape \cite{zins2021datadriven} and rigid objects~\cite{wang2021multi}. \sr{The fusion  layers are described in detail in the supplementary material.}

The encoder $E_S$ predicts the length of each bone. 
As the same human is analyzed along  all frames and views, the output is a single set of bone lengths. 

The encoder $E_Q$ predicts joint rotations, global root positions, and foot contact labels. 
Since 3D joint rotations and foot contact labels are identical to all views, $E_Q$ predicts a single set of rotations and contact labels per frame, shared by all views.
One exception is the root (pelvis) joint, whose rotation angle and position depend on the camera view and not on the human itself. Thus, for the root joint we predict the rotation angle and position for each frame and view. Root rotation and position are relative to the camera; hence visualizing the reconstructed object depicts the filmed person from the camera view, as expected.
Notice that root rotation and position carry the knowledge of the relative transformation between the cameras.
This insight suggests that our algorithm has the potential to output additional valuable information, \eg, cameras relative location to each other (left to future work).

At train time, the output of both encoders is combined in $K$ identical forward kinematic ($FK$) layers. 
Each $FK$ layer computes the estimated 3D joint positions related to one view, which in turn are 
 compared to the ground-truth for loss computation. 
In addition, temporal differences of the rotations extracted out of $E_Q$, are fed to a discriminator $D$~\cite{Kanazawa:2018}, so they get near the manifold of true rotations in an adversarial way. 


%
\sr{Formally,} let $\bL$ denote the number of bones, $\bT$ the temporal length of the sequence, $\bJ$ the number of joints, $\bQ$ the size of the rotations representation vector, and $\bK$ the number of cameras.
Let $\bP_{s,q,r}\in
\bbr
^{T\times 3J\times K}$ 
denote $K$ temporal sequences of 3D joint positions generated by a skeleton $\bss\in\bbr^L$ with joint rotations $\bq\in\bbr^{T\times Q\times(J-1)}$, and global root position and rotation $\br\in\bbr^{T\times (3+Q)\times K}$. Note that $q$ is related to all joints except for the root joint. The rotation of the root joint, as well as its position, are related to $r$. 

Our approach expects an input $\bV_{s,q,r}\in\bbr^{T\times 3J\times K}$ denoting $K$ temporal sequences of 2D joints and a confidence value per joint, 
related to a skeleton $s$, joint rotations $q$, and global root position and rotation $r$. 
Each input $V$ is fed into our deep neural network, which in turn predicts $\tilde{\bq}\in\mathbb{R}^{T\times Q\times(J-1)}$, that captures the dynamic, rotational information of the motion, $\tilde{\bss}\in\mathbb{R}^L$, that describes a single, consistent, skeleton, $\tilde{\br}\in\mathbb{R}^{T\times (3+Q)\times K}$ that estimates the global position and rotation of the root along time and along views, and $\tilde{\bff}\in\{0,1\}^{T\times 2}$ that predicts whether each of the two feet touches the ground in each frame:

\setlength{\abovedisplayskip}{-5pt plus 2pt minus 2pt}
\setlength{\belowdisplayskip}{5pt plus 2pt minus 2pt}

\begin{align}
\tilde{\bss} = E_S(F_S(\bV_{s,q,r})) ,  ~~~~~~~~~~
\tilde{\bq},\tilde{\br},\tilde{\bff} = E_Q(F_Q(\bV_{s,q,r})).
\end{align}
These attributes can be then combined via forward kinematics to estimate $K$ global 3D pose sequences, $\tilde{\bP}_{\tilde{s},\tilde{q},\tilde{r}}\in\mathbb{R}^{T\times 3J\times K}$, specified by joint positions:

\setlength{\abovedisplayskip}{-5pt plus 2pt minus 2pt}
\setlength{\belowdisplayskip}{5pt plus 2pt minus 2pt}

\begin{align}
\tilde{\bP}_{\tilde{s},\tilde{q},\tilde{r}}=FK(\tilde{\bss}, \tilde{\bq}, \tilde{\br}).
\end{align}



\iftog
\subsection{Losses}
\fi


We employ five loss functions. Our losses are inspired by Shi~\etal~\cite{shi2020motionet} and are enhanced to encompass the multitude of views.

\paragraphtinyvert{Joint Position Loss {\normalfont (the main loss)}} 
$\Loss_{\text{P}}$ ensures that joints in the extracted positions are in their correct 3D positions:

\setlength{\abovedisplayskip}{-7pt plus 2pt minus 2pt}
\setlength{\belowdisplayskip}{5pt plus 2pt minus 2pt}

\begin{align}
\Loss_{\text{P}} = \bbe_{\bP_{s,q,r}\sim \pp}\left[ \Vert FK(\tilde{\bss}, \tilde{\bq},\tilde{\br}_{pos_0})-\bP_{s,q,r_{pos_0}}\Vert^2 \right],
\end{align}

where $\bP_{s,q,r}\in \bbr^{T\times 3J\times K}$ denotes a 3D motion sequence, $\pp$ represents the distribution of 3D motion sequences in our dataset,
and $\tilde{r}_{pos_0}, r_{pos_0}$ stand for global position and rotation of the predicted and given root respectively, where the location is set to $(0,0,0)$, but the rotation is unchanged.

\paragraphtinyvert{Skeleton Loss}
$\Loss_{\text{S}}$ stimulates the skeleton branch of the network, $F_S$ and $E_S$, to correctly extract the skeleton $\bss$:
\ifeccv
    \setlength{\abovedisplayskip}{-7pt plus 2pt minus 2pt}
    \setlength{\belowdisplayskip}{5pt plus 2pt minus 2pt}
\fi

\begin{align}
\Loss_{\text{S}} = 
\bbe_{\bP_{s,q,r} \sim \pp} 
\left[ \Vert E_S(F_S(\bV_{s,q,r}))-\bss\Vert^2 \right].
\end{align}
\paragraphtinyvert{Adversarial Rotation Loss}
Our network learns to output rotations with natural velocity distribution using adversarial training. To achieve this, instead of focusing on rotation absolute values, like Kanazawa~\etal~\shortcite{Kanazawa:2018} we focus on the temporal differences of joint rotations. We create a discriminator $D_j$ for each joint. 
Note that the loss involving $D_{j\neq 0}$ takes the rotation values from $\tilde{q}$ while the loss involving $D_0$ takes the rotation values from $\tilde{r}$. It reads as
\setlength{\abovedisplayskip}{-5pt plus 2pt minus 2pt}
\setlength{\belowdisplayskip}{-5pt plus 2pt minus 2pt}

\begin{eqnarray*}
\Loss_{Q\_{GAN}_{j\ne 0}} \! & = \!&\bbe_{q \sim \qq}
\left[ \Vert D_j(\Delta_t q_j) \Vert^2 \right]+ \bbe_{\bP_{s,q,r}\sim \pp} \left[ \Vert 1-D_j(\Delta_t E_Q(F_Q((\bV_{s,q,r}))_{q_j}\Vert^2\right]
\end{eqnarray*}
%
\ifeccv
    \setlength{\abovedisplayskip}{0pt plus 2pt minus 2pt}
    \setlength{\belowdisplayskip}{5pt plus 2pt minus 2pt}
\fi
\begin{eqnarray}
\Loss_{{Q\_{GAN}}_{j= 0, k}} & =  &\ \bbe_{q \sim \qq}
\left[ \Vert D_j(\Delta_t q_j) \Vert^2 \right] \\ 
\nonumber
 &  + & 
 \ \bbe_{\bP_{s,q,r}\sim \pp} 
 \left[ \Vert 1-D_j(\Delta_t E_Q(F_Q((\bV_{s,q,r}))_{r_{{rot}_k}}\Vert^2\right],
\end{eqnarray}
where $\qq$ stands for the distribution of natural joint angles in the dataset, $E_Q(F_Q(\cdot))_{q_j}$ denotes the predicted rotations of the $j$th joint, ${{E_Q(F_Q(\cdot))_r}_{rot_k}}$ represents the predicted rotation of the pelvis joint relative to camera $k$, and $\Delta_t$ denotes temporal differences. 

\paragraphtinyvert{Global Root Position Loss}
We estimate the depth parameter, $Z_r\in\bbr^{T\times K}$, by minimizing:
\ifeccv
    \setlength{\abovedisplayskip}{0pt plus 2pt minus 2pt}
    \setlength{\belowdisplayskip}{7pt plus 2pt minus 2pt}
\fi
\begin{align}
\Loss_{\text{R}} = \bbe_{\bP_{s,q,r}\sim \pp}\left[ \Vert E_Q(F_Q(\bV_{s,q,r}))_{r_{pos_z}} - Z_r\Vert^2 \right],
\end{align}
where $Z_r$ is the depth of the ground-truth root, and  $E_Q(F_Q(\cdot))_{r_{pos_z}}$ is the depth of the predicted root. Note that $Z_r$ consists of values for all views and all frames.

\paragraphtinyvert{Foot Contact Loss}
We predict whether each foot contacts the ground in each frame and train the network via

\setlength{\abovedisplayskip}{-2pt plus 2pt minus 2pt}
\setlength{\belowdisplayskip}{7pt plus 2pt minus 2pt}

\begin{align}
\Loss_{\text{F}} = \bbe_{\bP_{s,q,r}\sim \pp}\left[ \Vert E_Q(F_Q(\bV_{s,q,r}))_f - \bff\Vert^2 \right],
\end{align}
where $E_Q(F_Q((\cdot))_f$ denotes the predicted foot contact label part ($\tilde{\bff}\in\{0,1\}^{T\times 2}$).
We encourage the velocity of foot positions to be zero during contact frames, by
\begin{align}
\Loss_{\text{FC}} = \bbe_{\bP_{s,q,r}\sim \pp}\left[ \Vert \bff_i \sum_j \Delta_t FK(\tilde{\bss}, \tilde{\bq}, \tilde{\br})_{f_i} \Vert^2 \right],
\end{align}
where $FK(\cdot,\cdot,\cdot)_{f_i}\in\mathbb{R}^{T\times 3}$ and $\bff_i$ denote the positions and the contact labels of one of the feet joints ($i\in \mathit{left}, \mathit{right}$), and $\sum_j$ sums the components for all axes.

Altogether, we obtain a total loss of:
\begin{eqnarray}
\Loss \!&=\!& \Loss_{\text{P}} +
\lambda_{\text{S}} \Loss_{\text{S}}
\lambda_{\text{Q}} \left(
\sum_{j\ne 0}\Loss_{{\text{Q\_GAN}}_j} +
\sum_{j=0,k}\Loss_{{\text{Q\_GAN}}_{j,k}}
\right) \\ \nonumber \!&+\!&
\lambda_{\text{R}}\Loss_{\text{R}} +
\lambda_{\text{F}}\Loss_{\text{P}_{F}} +
\lambda_{\text{FC}}\Loss_{\text{P}_{FC}}.
\end{eqnarray}
In most experiments we use $\lambda_{\text{S}}=0.1$, $\lambda_{\text{Q}}=1$, 
$\lambda_{\text{R}}=1.3$,  $\lambda_{\text{F}}=0.5$ and $\lambda_{\text{FC}}=0.5$. 

In the \ifarxiv{appendix  }\else{supplementary material }\fi we provide more implementation details, such as the description of each architectural block; in particular the novel fusion layers $F_S$ and $F_Q$. We discuss the advantages of early vs. middle and late fusion, and describe how we improve skeleton topology comparing to our single-view baseline. We also provide a detailed description of the datasets, a discussion of 2D pose estimators, and a description of the ground-truth we use.
\sectiontinyvert{Experiments and evaluation} \label{sec:experiments}


We present quantitative results on the Human3.6M~\cite{h36m_pami,IonescuSminchisescu11} and Ski-Pose PTZ-Camera~\cite{ski_ptz} datasets. We present qualitative results on the Human3.6M,  KTH Multi-view Football II~\cite{footballDS} and Ski-Pose PTZ-Camera~\cite{ski_ptz} datasets,
and on synthetic 
videos captured by dynamic cameras. Detailed description of these datasets can be found in the supplementary material.

\paragraphtinyvert{Quantitative results} 

\begin{table*}[t!]

\setlength{\abovecaptionskip}{5pt plus 3pt minus 2pt}
\setlength{\belowcaptionskip}{-2pt plus 3pt minus 2pt}

\caption{Protocol \#1 MPJPE error on Human3.6M. Legend:
$(*)$ is a \textbf{non} ep-free algorithm. In case parameters are not  given, we imitate their computation by perturbing the GT params by an unrealistically small perturbation amount; 
($\dagger$) exploit temporal information; ($+$) extra training data. In \bluebold{blue} - best result when camera parameters are not  given, in \textbf{bold} - best result per method group.
}

\resizebox{\textwidth}{!}{
\Large 
\begin{tabular}{c|ccccccccccccccc|c}
\toprule
\textbf{Method} &\textbf{Dir.} & \textbf{Disc.}&  \textbf{Eat} & \textbf{Greet} & \textbf{Phone} & \textbf{Photo} & \textbf{Pose} & \textbf{Purch.} &\textbf{Sit}& \textbf{SitD.}& \textbf{Smoke} &\textbf{Wait}& \textbf{WalkD.}& \textbf{Walk}& \textbf{WalkT.} & \textbf{Mean}\\
\midrule
\multicolumn{10}{l}{\textbf{Monocular} methods} & & & & & &\\
\midrule
Shi \etal~\shortcite{shi2020motionet}($\dagger$) &47.3 &53.1 &50.3 &53.9 &53.5 &52.8 &52.0 &55.4 &64.2 &54.8 &66.8 &55.0 &50.3 &59.1 &50.3 &54.6 \\
Llopart~\shortcite{llopart2020liftformer}($\dagger$) &42.2 &44.5 &42.6 &43.0 &46.9 &53.9 &42.5 &41.7 &55.2 &62.3 &44.9 &42.9 &45.3 &31.8 &31.8 &44.8 \\
Reddy \etal~\shortcite{Reddy2021TesseTrackEL}($\dagger$) &38.4 &46.2 &44.3 &43.2 &44.8 &48.3 &52.9 &\textbf{36.7} &\textbf{45.3} &54.5 &63.4 &44.4 &41.9 &46.2 &39.9 &44.6\\
{Li \etal~\shortcite{li2021exploiting}($\dagger$)} &39.9 &43.4 &40.0 &40.9 &46.4 &50.6 &42.1 &39.8 &55.8 &61.6 &44.9 &43.3 &44.9 &29.9 &30.3 &43.6 \\
{Hu \etal~\shortcite{hu2021conditional}($\dagger$)} &\textbf{35.5} &41.3 &\textbf{36.6} &39.1 &42.4 &49.0 &39.9 &37.0 &51.9 &63.3 &\textbf{40.9} &\textbf{41.4} &\textbf{40.3} &\textbf{29.8} &28.9 &41.1 \\
Cheng \etal~\shortcite{cheng20203d}~($\dagger$) &36.2 &\textbf{38.1} &42.7 &\textbf{35.9} &\textbf{38.2} &\textbf{45.7} &\textbf{36.8} &42.0 &45.9 &\textbf{51.3} &41.8 &41.5 &43.8 &33.1 &\textbf{28.6} &\textbf{40.1}  \\
\midrule
\midrule
\multicolumn{10}{l}{\textbf{Multi-view} methods, extrinsic camera parameters are \textbf{ given} } & & & & & &\\
\midrule
Tome \etal\shortcite{tome2018rethinking}~($+$) &43.3 &49.6 &42.0 &48.8 &51.1 &64.3 &40.3 &43.3 &66.0 &95.2 &50.2 &52.2 &51.1 &43.9 &45.3 &52.8 \\
\makecell{Kadkhodamohammadi \\ and  Padoy~\shortcite{kadkhodamohammadi2019generalizable}} &39.4 &46.9 &41.0 &42.7 &53.6 &54.8 &41.4 &50.0 &59.9 &78.8 &49.8 &46.2 &51.1 &40.5 &41.0 &49.1 \\
He \etal~\shortcite{he2020epipolar} &25.7 &27.7 &23.7 &24.8 &26.9 &31.4 &24.9 &26.5 &28.8 &31.7 &28.2 &26.4 &23.6 &28.3 &23.5 &26.9
\\
Qiu \etal~\shortcite{qiu2019cross}~($+$) &24.0 &26.7 &23.2 &24.3 &24.8 &22.8 &24.1 &28.6 &32.1 &26.9 &31.0 &25.6 &25.0 &28.0 &24.4 &26.2\\
Ma \etal~\cite{ma2021transfusion}($\dagger$) &24.4 &26.4 &23.4 &21.1 &25.2
&23.2 &24.7 &33.8 &29.8 &26.4 &26.8 &24.2 &23.2 &26.1 &23.3 &25.8\\
Iskakov \etal~\shortcite{iskakov2019learnable} &19.9 &20.0 &18.9 &18.5 &20.5 &19.4 &18.4 &22.1 &22.5 &28.7 &21.2 &20.8 &19.7 &22.1 &20.2 &20.8\\
Reddy \etal~\shortcite{Reddy2021TesseTrackEL}($\dagger$) &\textbf{17.5} &\textbf{19.6} &\textbf{17.2} &\textbf{18.3} &\textbf{18.2} &\textbf{17.7} &\textbf{18.0} &\textbf{18.0} &\textbf{20.5} &\textbf{20.3} &\textbf{19.4} &\textbf{17.2} &\textbf{18.9} &\textbf{19.0} &\textbf{17.8} &\textbf{18.7}\\
\midrule
\midrule
\multicolumn{10}{l}{\textbf{Multi-view} methods, extrinsic camera parameters are \textbf{not  given} } & & & & & &\\
\midrule
Chu and Pan~\shortcite{chu_and_pan_semisupervised}($\dagger$) &49.1 &63.6 &48.6 &56.0  &57.4 &69.6 &50.4 &62.0 &75.4 &77.4 &57.2 &53.5 &57.7 &37.6 &38.1 &56.9 \\
{\color{gray}
\makecell{Iskakov \etal~\shortcite{iskakov2019learnable}$(*)$ \\ param. perturb by 4\% }} &{\color{gray}30.2} &{\color{gray} 37.2} &{\color{gray} 32.7} &{\color{gray} 33.2} &{\color{gray} 38.8} &{\color{gray}43.7} &{\color{gray}29.7} &{\color{gray}43.0} &{\color{gray}49.4} &{\color{gray}67.6} &{\color{gray}38.0} &{\color{gray} 33.1} &{\color{gray}42.1} &{\color{gray}27.2} &{\color{gray}29.3} &{\color{gray}38.4} \\
{\color{gray}\makecell{Iskakov \etal~\shortcite{iskakov2019learnable}$(*)$ \\ param. perturb by 3\% }} &{\color{gray}27.6}  &{\color{gray}30.3}  &{\color{gray}29.0}  &{\color{gray}29.4}  &{\color{gray}33.1} &{\color{gray}{36.5}} 
&{\color{gray}27.4}  &{\color{gray}34.8}  &{\color{gray}39.1}  &{\color{gray}54.0} &{\color{gray}{34.4}}  &{\color{gray}30.7} &{\color{gray}36.2} &{\color{gray}{26.2}}  &{\color{gray}28.4}  &{\color{gray}33.1} \\
\textbf{Ours}($\dagger$) &\bluebold{22.0} &\bluebold{23.6} &\bluebold{24.9} &\bluebold{26.7} &\bluebold{30.6} &\bluebold{35.7} &\bluebold{25.1} &\bluebold{32.9} &\bluebold{29.5} &\bluebold{32.5} &\bluebold{32.6} &\bluebold{26.5} &\bluebold{34.7} &\bluebold{26.0} &\bluebold{27.7} &\bluebold{30.2}
\end{tabular}}

\label{tab:quant_human36}

\setlength{\abovecaptionskip}{-20pt plus 3pt minus 2pt}
\setlength{\belowcaptionskip}{-0pt plus 3pt minus 2pt}
\caption*{}

\end{table*}




We show quantitative results using the Mean Per Joint Position Error (MPJPE)~\cite{h36m_pami,IonescuSminchisescu11}, 
and report standard protocol \#1 MPJPE (that is, error relative to the pelvis), in millimeters.

\Cref{tab:quant_human36} presents a quantitative comparison of the MPJPE metric on the Human3.6M~\cite{h36m_pami} dataset. 
\sr{We present monocular methods, followed by multi-view ones that are split into ones that are acquainted with camera parameters and ones that are not.}
We show that in the absence of camera parameters, our model outperforms state-of-the-art methods by a large margin, and that even when camera parameters are available, 
FLEX is among the top methods. Note that these achievements are although FLEX aims at a slightly different task, which is motion reconstruction rather than pose estimation.

Being the only ep-free algorithm, we have no methods to compare to directly. However, algorithms can mitigate the lack of extrinsic camera parameters by estimating them. In the following comparisons, we show that when extrinsic parameters are not given, using estimated ones induces larger prediction errors, due to the innate inaccuracy of predicted values. On the other hand, FLEX is not affected by the lack of extrinsic parameters, since it does not use them whatsoever.
We compare FLEX with two models:
\begin{enumerate}[nosep,leftmargin=0cm,itemindent=0.5cm,labelwidth=\itemindent,labelsep=0cm,align=left]
\item[(1)]
There are two methods that do not use given camera parameters~\cite{chu_and_pan_semisupervised,kocabas2019selfsupervised}. 
They are not ep-free since they use estimated camera parameters, but we can still use them in settings where camera parameters are not given.
Only one of them~\cite{chu_and_pan_semisupervised} publishes MPJPE protocol \#1 results, and we significantly outperform it
(See \Cref{tab:quant_human36}). 
This gap is mostly because of the inaccuracy of parameter prediction and partially because their model is semi-supervised. 
    
\item[(2)] 
For comparing with the best available method, 
we have chosen the current state-of-the-art multi-view algorithm of Iskakov \etal~\cite{iskakov2019learnable} (TesseTrack~\cite{Reddy2021TesseTrackEL} is marginally better, but it does not provide code). Since their algorithm is \emph{not} ep-free, we
imitate parameter estimation by running a controlled perturbation of the camera parameters.
We re-train their method with distorted data to simulate an environment where camera distortion parameters are unknown. 
In addition, we perturb the 
extrinsic parameters by Gaussian noise with an extremely small standard deviation of 3\% of each parameter's 
value. That is, for a parameter $p$, we sample $\tilde{p}\sim\nn(p,(0.03 p)^2)$ and use $\tilde{p}$ as the input extrinsic parameter. 
We show that increasing the standard deviation from 3\% to 4\% yields a significant increase in the error, reflecting the sensitivity of non ep-free methods to  inaccuracy in camera parameters.
To obtain an equivalent environment, we compare FLEX to the method of Iskakov \etal~\shortcite{iskakov2019learnable} after using their own 2D pose estimation.
The lower part of \Cref{tab:quant_human36} shows that FLEX outperforms the non ep-free state-of-the-art, even when perturbation percentage is extremely small.
Their results
in that lower part 
are grayed out, to emphasize that we simulate an unrealistic setting.
\sr{Next, we show that a 3\% perturbation, rather than estimation, is fairer toward the compared method, as estimation induces larger inaccuracy.
%
We estimate the extrinsic camera parameters with two leading frameworks, COLMAP~\cite{schoenberger2016sfm} and OpenCV-SFM~\cite{opencv_library}, and obtain errors of 5.5\% and 8.6\%, respectively. The error is the mean value of $\frac{|p-\tilde{p}|}{p}$ for all extrinsic values $p$ and their estimation $\tilde{p}$. Moreover, the estimation process involves friction: OpenCV-SFM 
strongly depends on an initial guess, and COLMAP requires that each pair of cameras observes partially overlapping images, a limiting factor that prevents its usage in settings where the cameras face each other.}
\end{enumerate}



\begin{wraptable}{r}{0.48\textwidth}
\setlength{\abovecaptionskip}{-5pt plus 3pt minus 2pt}
\setlength{\belowcaptionskip}{-24pt plus 3pt minus 2pt}

\ifeccv
\caption{
MPJPE 
on the Ski-PTZ dataset,
measured for methods trained when extrinsic parameters are \emph{not} given. 
$(\dagger)$ is self/weakly-supervised. 
}
\fi
\begin{center}
\begin{tabular}{|c|c|}
\hline
\textbf{Method} & \textbf{MPJPE
} \\
\hline
CanonPose~\cite{wandt2020canonpose} ($\dagger$) & 128.1 \\
Chen \etal~\cite{chen2021deductive} ($\dagger$) & 99.4 \\
Ours & \textbf{65.5} \\
\hline
\end{tabular}
\end{center}

\ifeccv
\else
\caption{Protocol \#1 MPJPE error on the Ski-PTZ dataset,
measured for methods that are trained when extrinsic parameters are \textbf{not} given. 
Legend: $(\dagger)$ is self-supervised. 
$(*)$ is weakly supervised.}
\fi

\label{tab:ski_quantitative}

\setlength{\abovecaptionskip}{-50pt plus 3pt minus 2pt}
\setlength{\belowcaptionskip}{-0pt plus 3pt minus 2pt}
\caption*{}

\end{wraptable}

In addition to the comprehensive comparison on the Human3.6 dataset, 
in \Cref{tab:ski_quantitative} we show a quantitative comparison on the Ski-Pose PTZ-Camera~\cite{ski_ptz} dataset, for methods that are trained when camera parameters are \emph{not} given.
These methods are comparable in settings that lack extrinsic parameters because they estimate them. 
However, since they still use (estimated) parameters, they are not ep-free.
FLEX leads the table with a large gap.
This gap is mostly because parameter estimation induces an inevitable inaccuracy, and partially because the compared models are self/semi-supervised. 

\begin{table}[t!]
\setlength{\abovecaptionskip}{-5pt plus 3pt minus 2pt}
\setlength{\belowcaptionskip}{-5pt plus 3pt minus 2pt}

    \hfill
    \begin{minipage}{.35\linewidth}
        \centering
        \caption{Smoothness, measured by acceleration error ($mm/s^2$), on Human3.6M.\\
        ($\star$):~2D pose from \cite{iskakov2019learnable}.\hfill\\ ($\bullet$):~ground-truth 2D poses.}
        \begin{center}
\begin{tabular}{|c|c|}
\hline
\ifeccv \else \kern-3pt \fi
\textbf{Method} & \textbf{Acc. Err.}$\ \downarrow$ \\ \hline
VIBE\cite{kocabas2020vibe} & 18.3 \\ \hline
MEVA\cite{luo20203d} & 15.3 \\ \hline
HMMR\cite{kanazawa2019learning} & 9.1 \\ \hline
TCMR\cite{choi2021static} & 5.3 \\ \hline
Iskakov\cite{iskakov2019learnable} & 3.9\\ \hline
Shi\cite{shi2020motionet}  & 3.6($\star$) / 2.0($\bullet$) \\ \hline
FLEX  & \textbf{1.6}($\star$) / \textbf{0.9}($\bullet$) \\ \hline

\end{tabular}
\end{center}



        \ifeccv
        \else
        \caption{Smoothness, measured by acceleration error ($mm/s^2$), on Human3.6M.\\
        ($\star$):~2D pose from \cite{iskakov2019learnable}.\hfill\\ ($\bullet$):~ground-truth 2D poses.}
        \fi
        \label{tab:acc_error}
    \end{minipage}%
    \hfill
    \begin{minipage}{.36\linewidth}
      \centering
        \caption{Attention impact.\\ TE: Transformer Encoder.\hfill \\MHA: Multi-head Attention.\hfill\\ $l$: no. of stacked layers.\\ $h$: no. of attention heads.}
\begin{center}
\begin{tabular}{|c|c|}
\hline

\textbf{Method} & \textbf{MPJPE} \\

\hline
Conv. layer & 31.9 \\
\hline
TE - 1$l$, 64$h$ & 30.9 \\
\hline
TE - 2$l$, 64$h$ & 37.8 \\
\hline
MHA - 128$h$ & 30.5 \\
\hline
MHA - 64$h$ & \textbf{30.2} \\
\hline
MHA - 32$h$ & 30.6 \\
\hline
MHA - 16$h$ & 30.9 \\
\hline

\end{tabular}
\end{center}

        \label{tab:fusion_arch}
    \end{minipage}
    \hfill
    \hphantom{.} 
    
\setlength{\abovecaptionskip}{-35pt plus 3pt minus 2pt}
\setlength{\belowcaptionskip}{-0pt plus 3pt minus 2pt}
\caption*{}

\end{table}

A known strength of predicting rotation angles rather than locations, is the \emph{smoothness} of predicted motion. 
In \Cref{tab:acc_error} we show that FLEX's smoothness result outperforms others by a large margin. Following Kanazawa \etal~\cite{kanazawa2019learning}, we measure smoothness using the acceleration error of each joint. 


\paragraphtinyvert{Qualitative results}

\begin{figure}[t]
\setlength{\abovecaptionskip}{5pt plus 3pt minus 2pt}
\setlength{\belowcaptionskip}{-5pt plus 3pt minus 2pt}
\centering
\begin{minipage}[b]{0.48\linewidth}
    \centering
    \includegraphics[width=\textwidth]{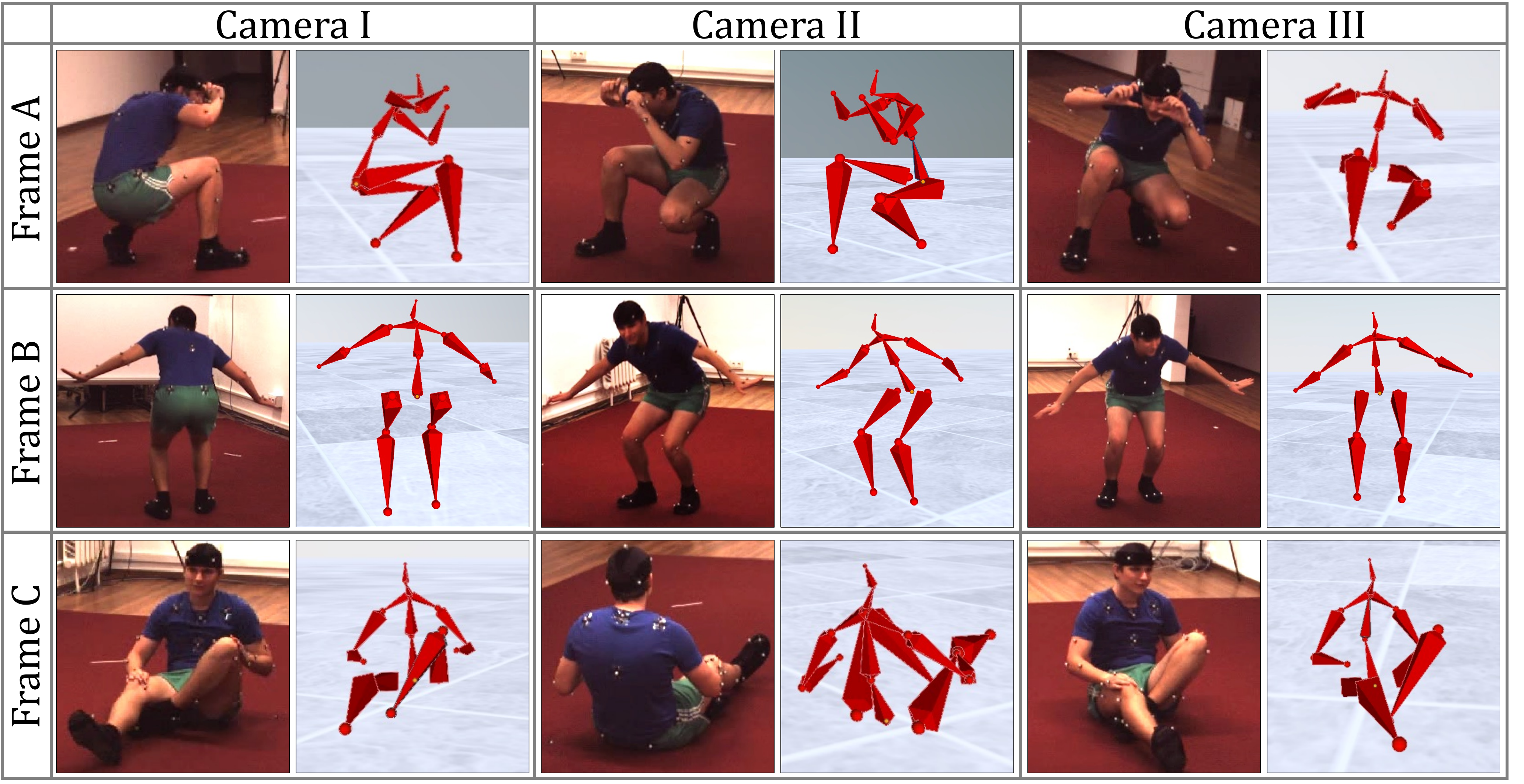}
    \caption{Our results on videos from the Human3.6M dataset.}
    \label{fig:quality_h36}
\end{minipage}
\hfill
\begin{minipage}[b]{0.48\linewidth}
    \centering
    \includegraphics[width=\textwidth]{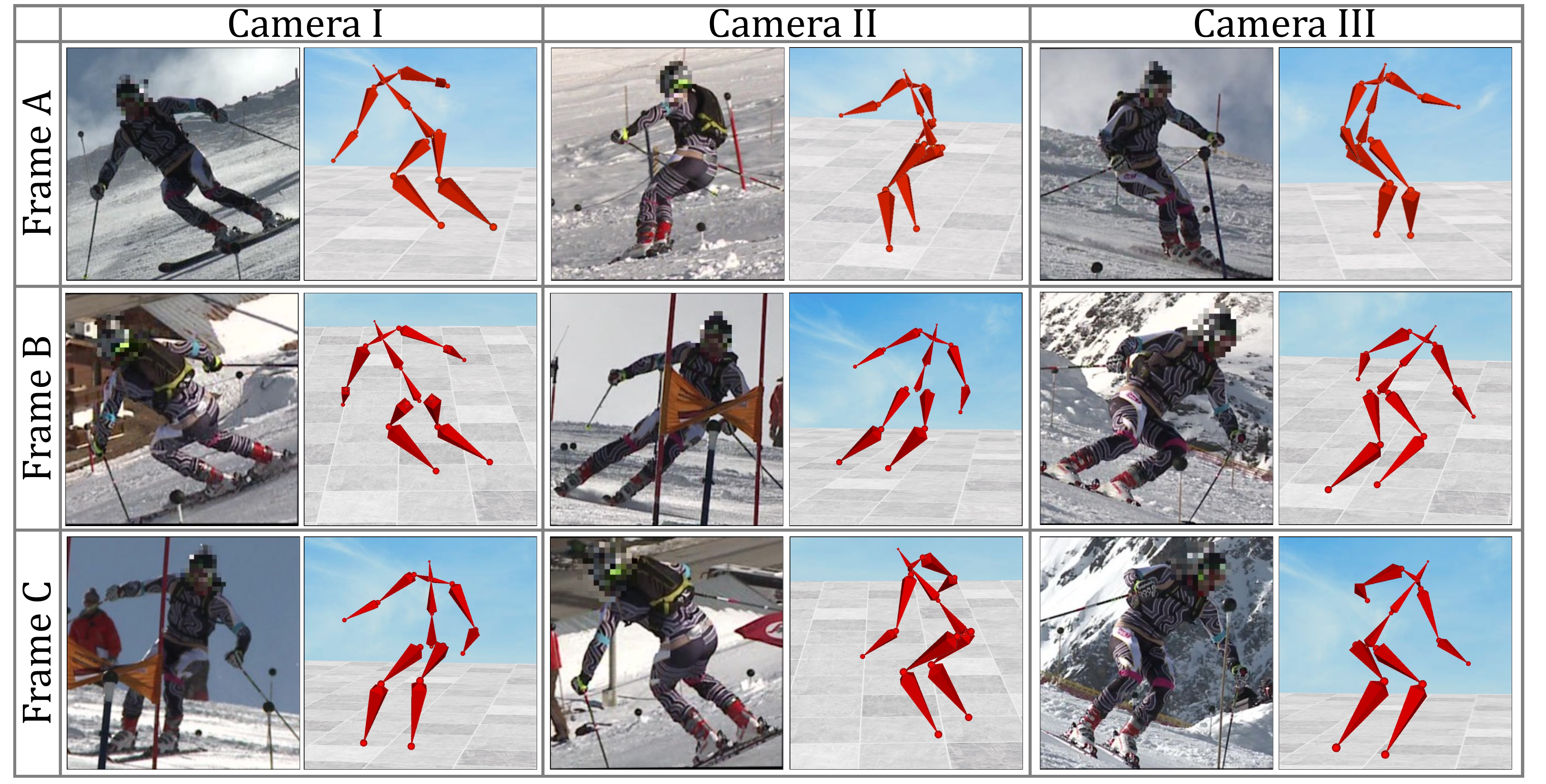}
    \caption{Our results on videos from the Ski-Pose PTZ-Camera dataset.}
    \label{fig:ski_ptz_qualitative}
\end{minipage}
\end{figure}

In the following figures we show rigs, that is, bone structure 
from reconstructed animation videos, selecting challenging 
scenes. Videos of the reconstructed motions are  
\ifanonymous{provided \iftog{as well,}\else{in the sup. mat.,}\fi}
\else{available on our project page,}
\fi
presenting the smoothness of motion and the naturalness of rotations. 
\Cref{fig:football_teaser,fig:quality_h36,fig:ski_ptz_qualitative}
 show scenes from the KTH Multi-view Football II~\cite{footballDS}, the Human3.6M~\cite{h36m_pami,IonescuSminchisescu11} and the Ski-Pose PTZ-Camera~\cite{ski_ptz} datasets, respectively. 
Each row depicts three views of one time frame. To the right of each image, we place a reconstructed rig, which is sometimes zoomed in for better visualization. 
Notice the occluded and blurry scenes in the football figure (\ref{fig:football_teaser}). The KTH Football dataset is filmed using dynamic (moving) cameras, 
a setting where extrinsic parameters are rarely given, thus disqualifying methods that require camera parameters.
Our algorithm is agnostic to the lack of camera parameters and attains good qualitative results. 

\begin{figure}[t]
\setlength{\abovecaptionskip}{5pt plus 3pt minus 2pt}
\setlength{\belowcaptionskip}{-15pt plus 3pt minus 2pt}
\centering
\begin{minipage}[b]{0.68\linewidth}
    \centering
    \includegraphics[width=\textwidth, height=5.0cm]{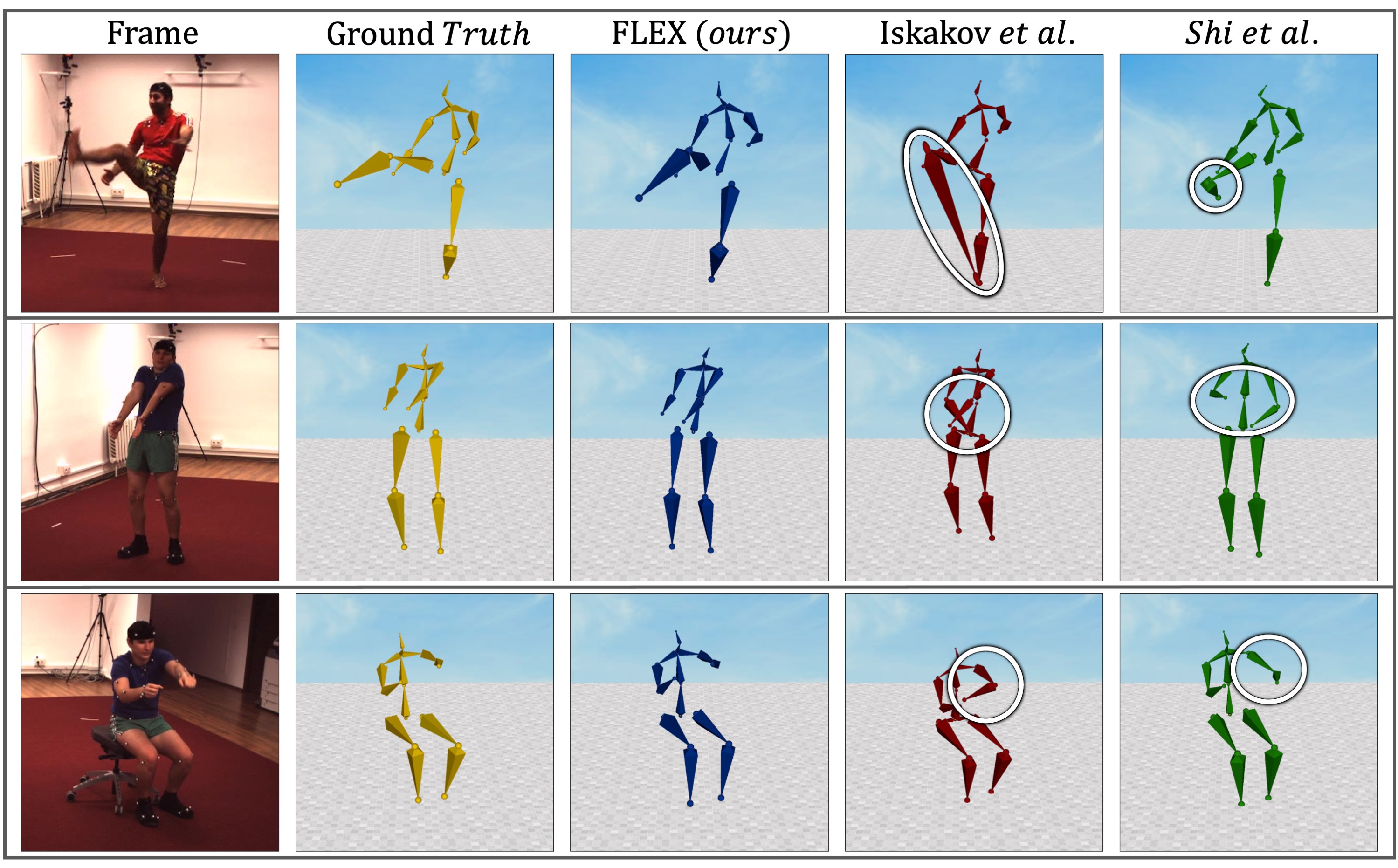}
    \caption{Qualitative comparison of our work vs. non ep-free state-of-the-art (Iskakov \etal~\cite{iskakov2019learnable}) and vs. our single-view baseline (Shi \etal~\cite{shi2020motionet}).}
    \label{fig:competitors}
\end{minipage}
\hfill
\begin{minipage}[b]{0.27\linewidth}
    \centering
    \includegraphics[width=0.8\textwidth,height=5.0cm]{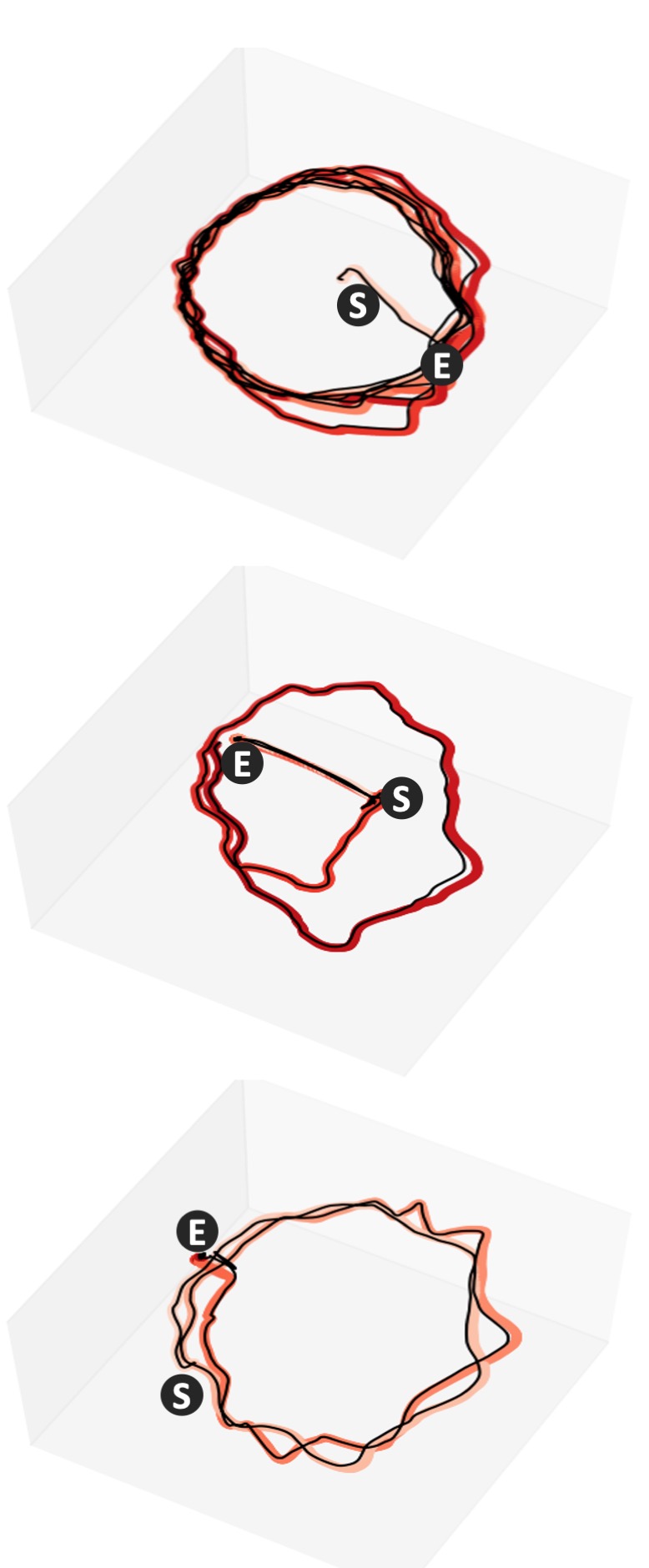}
    \caption{Global root position. Ground-truth is in thin black.}
    \label{fig:root_pos}
\end{minipage}
\end{figure}

In \Cref{fig:competitors} we show qualitative results of FLEX, compared to current non ep-free  multi-view state-of-the-art \cite{iskakov2019learnable}, and to our monocular baseline~\cite{shi2020motionet}. 
Note that the method in \cite{iskakov2019learnable} produces unnatural poses such as a huge leg in the first row and a backward-bent elbow in the last row. 

\begin{figure*}[htb]
\setlength{\abovecaptionskip}{5pt plus 3pt minus 2pt}
\setlength{\belowcaptionskip}{-15pt plus 3pt minus 2pt}
\centering
\includegraphics[width=\linewidth]{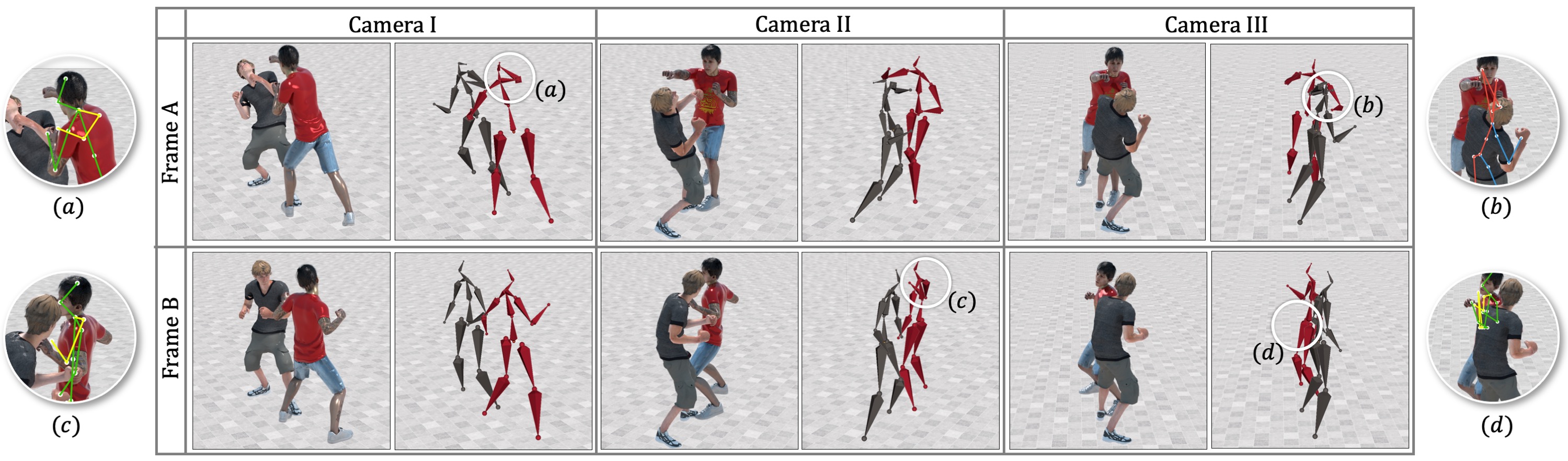}
\caption{Results on multi-person synthetic videos.
\sr{In the zoomed-in circular images we depict 2D pose estimations, which are erroneous due to occlusion. A matching circle in the center rectangular image shows that 
our method reconstructs correct 3D motion although it takes inaccurate 2D joints for input. 
}
} 
\label{fig:fight_results_with_circles}
\end{figure*}

\paragraphtinyvert{Multi-person captured by dynamic cameras} 
We evaluate our algorithm on
a setting with dynamic cameras, with multi-person scenes introducing
severe inter-person occlusions.
Recall that the term \emph{dynamic} refers to moving cameras that occasionally change their location and rotation.
There are several multi-view datasets. Most of them are not fully dynamic: Human3.6M~\cite{h36m_pami,IonescuSminchisescu11}, CMU Panoptic~\cite{CMU:mocap} and TUM Shelf $\&$ Campus~\cite{campus_shelf} contain static scenes only, while Tagging~\cite{tagging_dataset} and Ski-Pose PTZ-Camera~\cite{ski_ptz} contain  rotating cameras whose locations are fixed. KTH~\cite{footballDS} is fully dynamic, but it is too blurry and does not provide ground-truth for all subjects.
Despite its limitations, we use the KTH dataset for qualitative analysis, but we cannot use it for thorough research.
To mitigate the lack of a dynamic dataset, we generate synthetic videos using animated characters downloaded from Mixamo~\cite{mixamo}, an online dataset of character animation. Then, we generate video sequences of two interacting characters using Blender~\cite{blender}, which is a 3D creation suite. The newly created data is available on our project page.
Our "synthetic studio" is illustrated at the \ifarxiv{appendix}\else{sup. mat.}\fi, where two interacting figures are video-filmed by multiple dynamic cameras.
Using Blender, we obtain a rendered video stream from the view angle of each synthetic camera. 
Recall that the input to our algorithm is 2D joint locations, hence it is agnostic to the video appearance, and to whether the input image is real or synthetic.

The 2D backbone we use over the rendered video sequences is Alphapose~\cite{alphapose}, a state-of-the-art multi-person 2D pose estimator.
Once obtaining the 2D joint locations, we use a na\"ive heuristic, which is not part of the suggested algorithm, to associate each detected person with its ID: for each frame, we associate the detected 2D pose with the one that is geometrically closest to it in the previous frame. 
In \Cref{fig:fight_results_with_circles} we depict qualitative results of two boxers. We emphasize several viewpoints where the 2D estimator attains large errors. Yet, FLEX compensates for these errors by fusing multi-view information. In the \ifarxiv{appendix }\else{sup. mat. }\fi we show additional characters and the predicted 2D pose for all the viewpoints.

\paragraphtinyvert{Global position}

In \Cref{fig:root_pos} we draw the global position of the scaled predicted root joint along time.
Ground-truth is depicted using a thin black curve, and our prediction is an overlay on top of it, changing from light to dark as time progresses. The start and the end of each trajectory are signaled by the letters S and E, respectively.
Depicted motions are evaluated on the test set of Human3.6M, on the motions of walking, talking on the phone, and eating.
Note that our predictions almost completely overlap the ground-truth curve. 
Recall we use weak perspective to bypass dependency on intrinsic parameters, resulting in up-to-scale global position accuracy. 
\sr{Quantitatively, our MPJPE on the H36M validation
set is $118mm$, outperforming Iskakov \etal~\cite{iskakov2019learnable} (perturbed by 3\%) that attain $123mm$. The other ep-free work \cite{chu_and_pan_semisupervised} does not solve global locations.}


\paragraphtinyvert{Ablation study}

\ifeccv{
    \begin{table}[t]
\setlength{\abovecaptionskip}{0pt plus 3pt minus 2pt}
\setlength{\belowcaptionskip}{10pt plus 3pt minus 2pt}
    \caption{Ablation studies: The impact of (a) Number of views; (b) 2D backbone, and (c) Fusion method \sr{(refer to the sup. mat. for details regarding fusion)}.}
    \begin{minipage}{.32\linewidth}
\setlength{\abovecaptionskip}{0pt plus 3pt minus 2pt}
\setlength{\belowcaptionskip}{-10pt plus 3pt minus 2pt}
        \centering
        \caption*{\small(a)}


\begin{tabular}{|c|c|c|}
\hline
\multirow{2}{*}{\textbf{\#Views}} & \multicolumn{2}{c|}{\textbf{2D backbone}} \\ \cline{2-3} 
 & \rule{0pt}{2ex}\textbf{GT } & \textbf{\cite{iskakov2019learnable}}\\
\hline
1 & 47.7 & 56.3 \\
\hline
2 & 33.9 & 41.4\\
\hline
3 & 26.3 & 34.6 \\
\hline
4 & \textbf{22.9} & \textbf{30.2} \\
\hline
\end{tabular}


        \label{tab:num_views}
    \end{minipage}
    \hfill
    \begin{minipage}{.32\linewidth}
\setlength{\abovecaptionskip}{-10pt plus 3pt minus 2pt}
\setlength{\belowcaptionskip}{0pt plus 3pt minus 2pt}
        \centering
        \caption*{\small(b)}



\begin{center}

\begin{tabular}{|c|c|}
\hline 
\ifeccv \else \kern-3pt \fi
\makecell{\textbf{2D} \\ 
\textbf{backbone}} & 
\ifeccv \else \kern-3pt \fi 
\textbf{MPJPE }
\ifeccv \else \kern-3pt \fi
\\
\hline
\cite{Cao:2018} &38.6\\
\hline
\cite{chen2018cascaded} &31.7\\
\hline
\cite{iskakov2019learnable} &30.2\\
\hline
GT &\textbf{22.9}\\
\hline

\end{tabular}
\end{center}

        \label{tab:2D_backbone}
    \end{minipage}
    \hfill
    \begin{minipage}{.32\linewidth}
\setlength{\abovecaptionskip}{-11pt plus 3pt minus 2pt}
\setlength{\belowcaptionskip}{-21pt plus 3pt minus 2pt}
        \centering
        \caption*{\small(c)}
        \begin{center}
\begin{tabular}{|c|c|}
\hline
\textbf{Method} & \textbf{MPJPE } \\
\hline
Averaged $K$ views & 36.4 \\
\hline
Late fusion & 31.0 \\
\hline
FLEX & \textbf{22.9} \\
\hline
\end{tabular}
\end{center}
        \label{tab:ablation}
    \end{minipage}
\setlength{\abovecaptionskip}{-35pt plus 3pt minus 2pt}
\setlength{\belowcaptionskip}{-0pt plus 3pt minus 2pt}
\caption*{}
\end{table}

}
\else{
    \begin{table}[t]
    \begin{minipage}{.55\linewidth}
      \centering
      \ifeccv
      \caption{No. of views impact.}
      \fi


\begin{tabular}{|c|c|c|}
\hline
\multirow{2}{*}{\textbf{\#Views}} & \multicolumn{2}{c|}{\textbf{2D backbone}} \\ \cline{2-3} 
 & \rule{0pt}{2ex}\textbf{GT } & \textbf{\cite{iskakov2019learnable}}\\
\hline
1 & 47.7 & 56.3 \\
\hline
2 & 33.9 & 41.4\\
\hline
3 & 26.3 & 34.6 \\
\hline
4 & \textbf{22.9} & \textbf{30.2} \\
\hline
\end{tabular}


        \label{tab:num_views}
    \end{minipage}%
    \begin{minipage}{.45\linewidth}
      \centering
      \caption{2D backbone impact.}



\begin{center}

\begin{tabular}{|c|c|}
\hline 
\ifeccv \else \kern-3pt \fi
\makecell{\textbf{2D} \\ 
\textbf{backbone}} & 
\ifeccv \else \kern-3pt \fi 
\textbf{MPJPE }
\ifeccv \else \kern-3pt \fi
\\
\hline
\cite{Cao:2018} &38.6\\
\hline
\cite{chen2018cascaded} &31.7\\
\hline
\cite{iskakov2019learnable} &30.2\\
\hline
GT &\textbf{22.9}\\
\hline

\end{tabular}
\end{center}

       \label{tab:2D_backbone}
    \end{minipage} 
\end{table}
    \begin{table}[tb]
\ifeccv
\caption{The impact of various fusion methods with 2D GT input. Note that processing all views together with an early fusion as done in FLEX outperforms the other variations by a large margin.}
\fi
\begin{center}
\begin{tabular}{|c|c|}
\hline
\textbf{Method} & \textbf{MPJPE } \\
\hline
Averaged $K$ views & 36.4 \\
\hline
Late fusion & 31.0 \\
\hline
FLEX & \textbf{22.9} \\
\hline
\end{tabular}
\end{center}
\ifeccv
\else
\vspace{-15pt}
\caption{The impact of various fusion methods with 2D GT input. Note that processing all views together with an early fusion as done in FLEX outperforms the other variations by a large margin.}
\fi
\label{tab:ablation}
\vspace{-10pt}
\end{table}

}
\fi

We evaluate the impact of different settings on the performance of FLEX using various ablation tests. 
\Cref{tab:fusion_arch} compares different multi-view fusion architectures.  
Note that using attention rather than convolution yields a 2mm improvement.
The performance degrades with the transformer encoder due to its large number of parameters, which require more data for training than what is available in our case.

\Cref{tab:num_views} measures MPJPE on Human3.6M in several studies.
\Cref{tab:num_views}(a) studies a varying number of views, where the 2D pose is once given and once estimated. It confirms that a larger number of views induces more accurate results. Note that the gap between the two columns decreases once the number of views increases. It shows that using several views compensates for the inaccuracy of estimated 2D poses. 
\Cref{tab:2D_backbone}(b) compares 2D pose estimation backbones, and justifies our use of Iskakov \etal~\cite{iskakov2019learnable}.
%
Finally, in \Cref{tab:ablation}(c) we explore
two variations, both with ground-truth 2D inputs. 
The first variation runs FLEX as a monocular method (K=1) and averages the monocular predictions.
%
The second changes the fusion layers, $F_S$ and $F_Q$, to use late fusion instead of an early one. 
We conclude that the configuration used by FLEX is better than both variations.

\begin{wrapfigure}{R}{0.5\textwidth}
\centering

\setlength{\abovecaptionskip}{-32pt plus 3pt minus 2pt}
\setlength{\belowcaptionskip}{0pt plus 3pt minus 2pt}
\caption*{}
\includegraphics[width=0.5\textwidth]{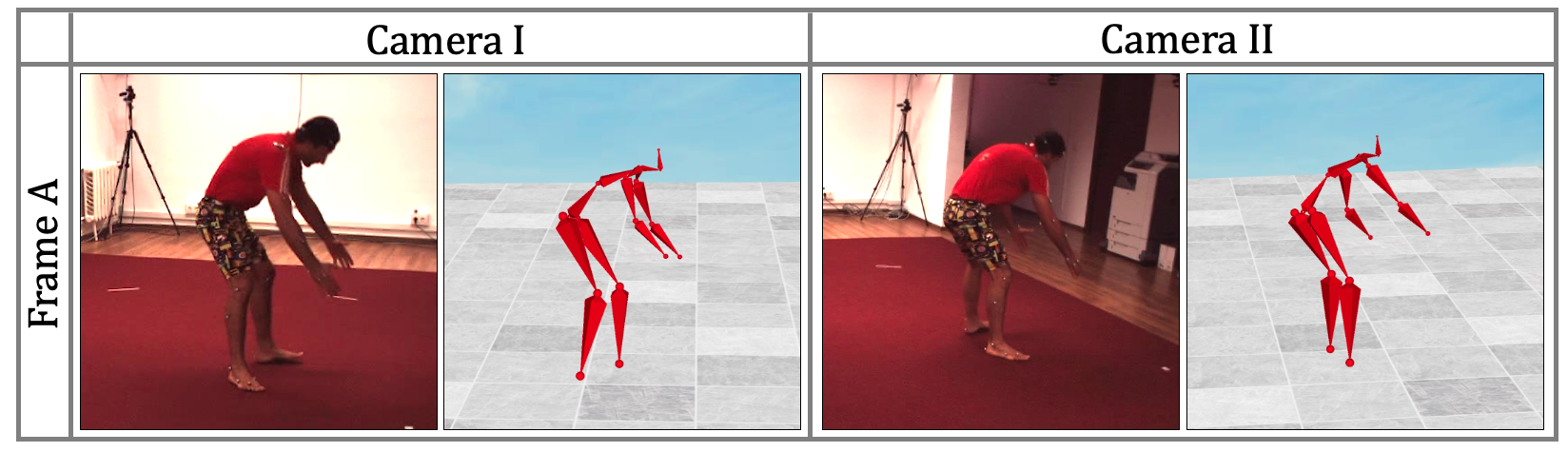}

\setlength{\abovecaptionskip}{-33pt plus 3pt minus 2pt}
\setlength{\belowcaptionskip}{0pt plus 3pt minus 2pt}
\caption*{}
\label{fig:generalization_inset}

\end{wrapfigure}

\setcolor{violet}
\paragraphtinyvert{Generalization}
%
%
We exhibit generalization by training on one dataset and evaluating on a different, more challenging one. The train dataset is Human3.6M, and the evaluation ones are the KTH Football dataset, and the synthetic videos.
%
For quantitative measurement, we train our model on two of the four cameras of the Human3.6M dataset.
We test it using the other two cameras, on which the model has not been trained. 
We repeat this process for all possible camera pairs and obtain an average MPJPE of 148mm. Note that this error is not large compared to the human body size, and indeed we attain pleasing visual results as shown in the inset on the right.
\setcolor{black}

\sectiontinyvert{Conclusions and limitations}

We have presented FLEX, a multi-view method for motion reconstruction. 
It relies on a key understanding that 3D rotation angles and bone lengths are invariant to camera view, and their direct reconstruction spares the need for camera parameters. 
On a technical viewpoint, we presented a novel fusion layer with a multi-view convolutional layer and a multi-head attention mechanism that attends views.

One limitation of our approach is the dependency on 3D joint location ground-truth, and in particular, the requirement that it is given at the axis system of the train cameras. 
Another limitation is the dependency on the 2D backbone quality, and on the accuracy value associated with each joint.
Lastly, being ep-free, the output 3D joint positions are only relative to the camera, lacking the transformation with respect to a global axis system.  

In summary, FLEX is unique in fusing multi-view information to reconstruct motion and pose in dynamic photography environments. %
It is unaffected by settings in which the relative rotations between the cameras are unknown, 
and can maintain a high level of accuracy regardless.
FLEX offers a simpler setting, where the correspondence and compatibility among the different views are rather lean, and thus more resilient to input errors and innate inaccuracies.

\ifanonymous{}\else{
\subsubsectiontinyvert{Acknowledgments}
This research
would not have been possible without the exceptional support of Mingyi Shi. 
We are grateful to Kfir Aberman and Yuval Alaluf for reviewing earlier versions of the manuscript, and to Yuval Alaluf and Shahaf Goren for contributing to FLEX's video clip.
This work was supported in part by the Israel Science Foundation (grants no. 2366/16 and 2492/20).
}\fi

\bibliographystyle{ECCV22/splncs04}
\bibliography{main}

\ifarxiv{
\newpage
\appendix

\ifarxiv{
\section*{Appendix}
}
\else{
\sectiontinyvert{Outline}
}
\fi

\ifarxiv{
This appendix provides further details for the main paper and is constructed in the following way.
}
\else{
The following document is constructed in the following way.
}
\fi
 \Cref{sec:code} describes how to access our model's code, and \Cref{sec:media} provides additional media, namely figures that have not been included in the main paper due to page limitation, and a description of video files that are \ifanonymous{given as part of the sup. mat}\else{available at our project page\footnotemark[1]{}\footnotetext[1]{Project page: \projectpage}}\fi. In \Cref{sec:skeleton} we describe an improvement in the skeleton structure, and in \Cref{sec:architecture_details} we detail the internals of our architecture. \Cref{sec:data} thoroughly describes the datasets and various data aspects of our work, and finally,  \Cref{sec:cam_param_technical} presents technical details related to camera parameters.

\ifeccv
\vspace{18pt}
\fi

\sectiontinyvert{Code} \label{sec:code}
Our code, together with usage instructions, 
\ifanonymous{is given as part of the sup. mat. It can also be found in our anonymous project page\footnote{\label{footnote_github}\url{https://github.com/FLEX-2021/FLEX}.}
}
\else{is available on our project page\footnotemark[1]{}.}
\fi
The reader is encouraged to run the code and witness the reproducibility of our model.

\sectiontinyvert{Additional visualizations} \label{sec:media}

\ifanonymous{Together with this document, we provide supplementary video files.}
\else{On our project page\footnotemark[1], the reader can find attached video files.}
\fi 
The reader is encouraged to browse the video files in full-screen size. 
\ifanonymous{
A higher quality version of these files can be found in our anonymous project page\footnotemark[1]. 
}
\fi
Here is their description:

\begin{itemize}
    \item A clip describing our work: clip.mp4
    \item Video files showing our results on the Human3.6M dataset: Human36M*.mp4
    \item Video files showing our results on the KTH multi-view Football II dataset: KTH\_football.mp4
    \item Video files comparing MotioNet (single-view) and Iskakov \etal~\cite{iskakov2019learnable} results versus ours: MotioNet\_comparison.mp4 and Iskakov\_comparison.mp4, respectively. 
\end{itemize}

Note that we present only results that use input obtained by 2D estimation (as opposed to ground truth). Thus, our input is affected by occlusion and blur. Yet, we are able to mitigate the noisy input by exploiting multi-view data, in an ep-free fashion.

\begin{figure}[tbh]
\centering
\includegraphics[width=\linewidth]{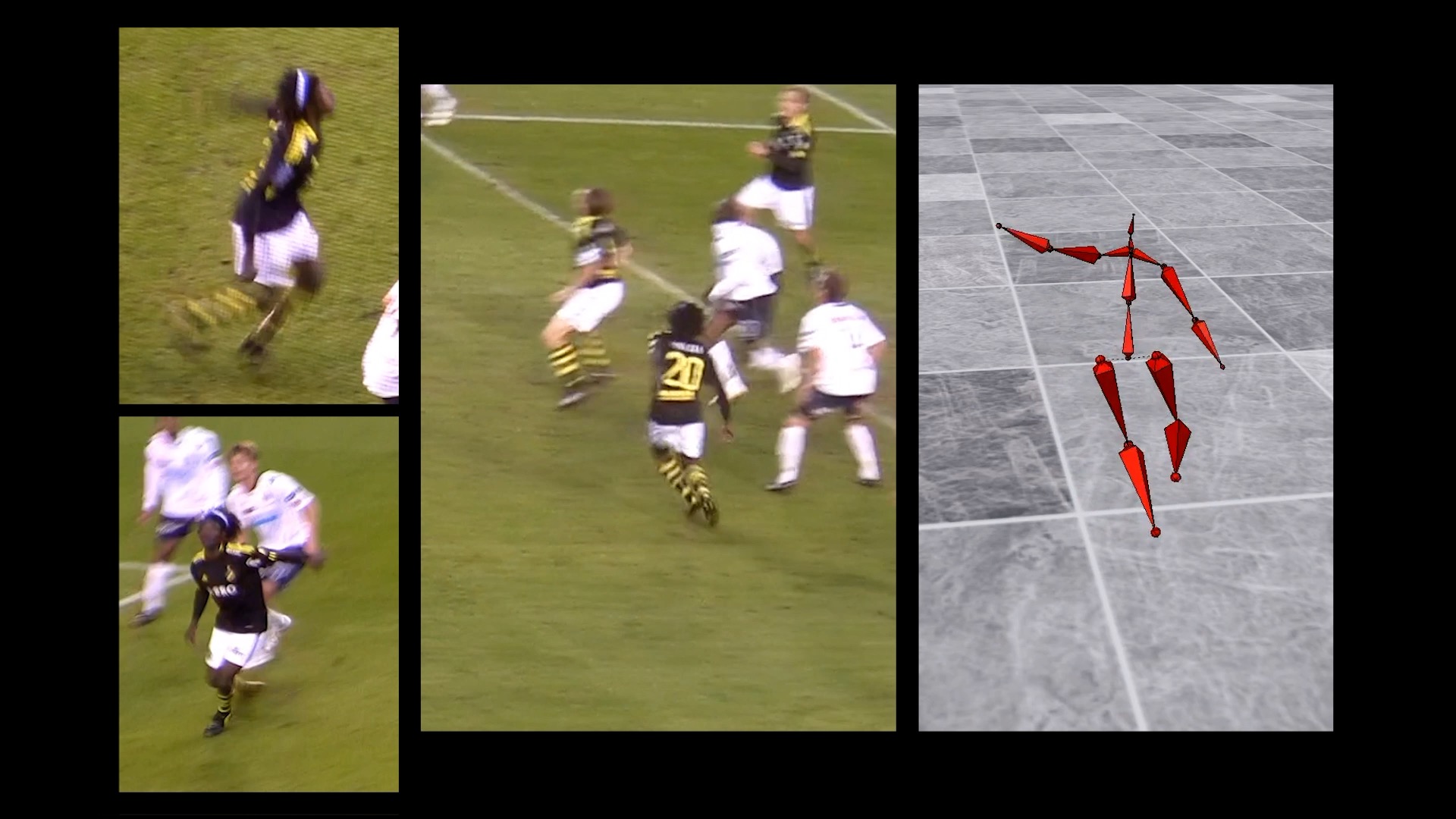}
\setlength{\abovecaptionskip}{-5pt plus 3pt minus 2pt}
\setlength{\belowcaptionskip}{-5pt plus 3pt minus 2pt}
\caption{Our algorithm is able to grasp fine details. The player's left hand cannot be seen in the center view and is blurred in the left views. Yet, our model accurately reconstructs it. }
\label{fig:quality_football_hand_wave}
\end{figure}

In \Cref{fig:quality_football_hand_wave} we show how our algorithm is able to grasp fine details. The player's left hand cannot be seen in the center view and is blurred in the left views. Yet, our model accurately reconstructs it.

In \Cref{fig:quality_h36_b,fig:quality_KTH_b}, we show additional results on the Human3.6M and KTH Football multi-view II datasets. Each row depicts three views of one time frame. To the right of each image we place a reconstructed rig. 
\ifeccv{
\Cref{fig:quality_ski_large,fig:quality_h36_large,fig:quality_KTH_large} are enlarged versions of the figures shown in the main paper.
}\fi

\begin{figure}[tbh]
\centering
\includegraphics[width=\linewidth]{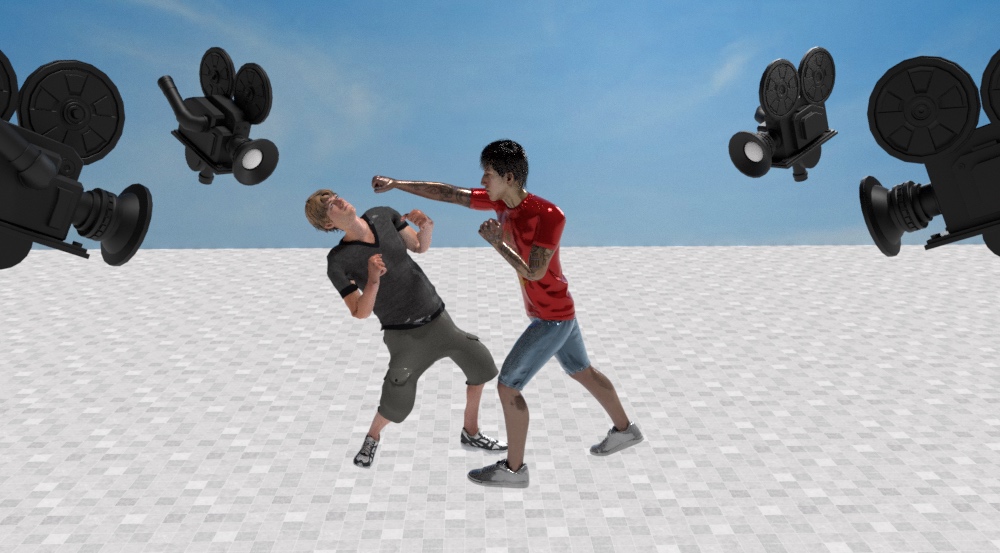}
\setlength{\abovecaptionskip}{-5pt plus 3pt minus 2pt}
\setlength{\belowcaptionskip}{-5pt plus 3pt minus 2pt}
\caption{Our "synthetic studio" created using Blender~\cite{blender} software with Mixamo~\cite{mixamo} 3D characters.
Two interacting characters are captured by multiple dynamic cameras and rendered into multiple video streams.}
\label{fig:blender_studio}
\end{figure}
\Cref{fig:blender_studio} shows our recording setup for creation of synthetic data. Note the depicted cameras, that dynamically move in the scene.

\begin{figure*}[tbh]
\centering
\includegraphics[width=\linewidth]{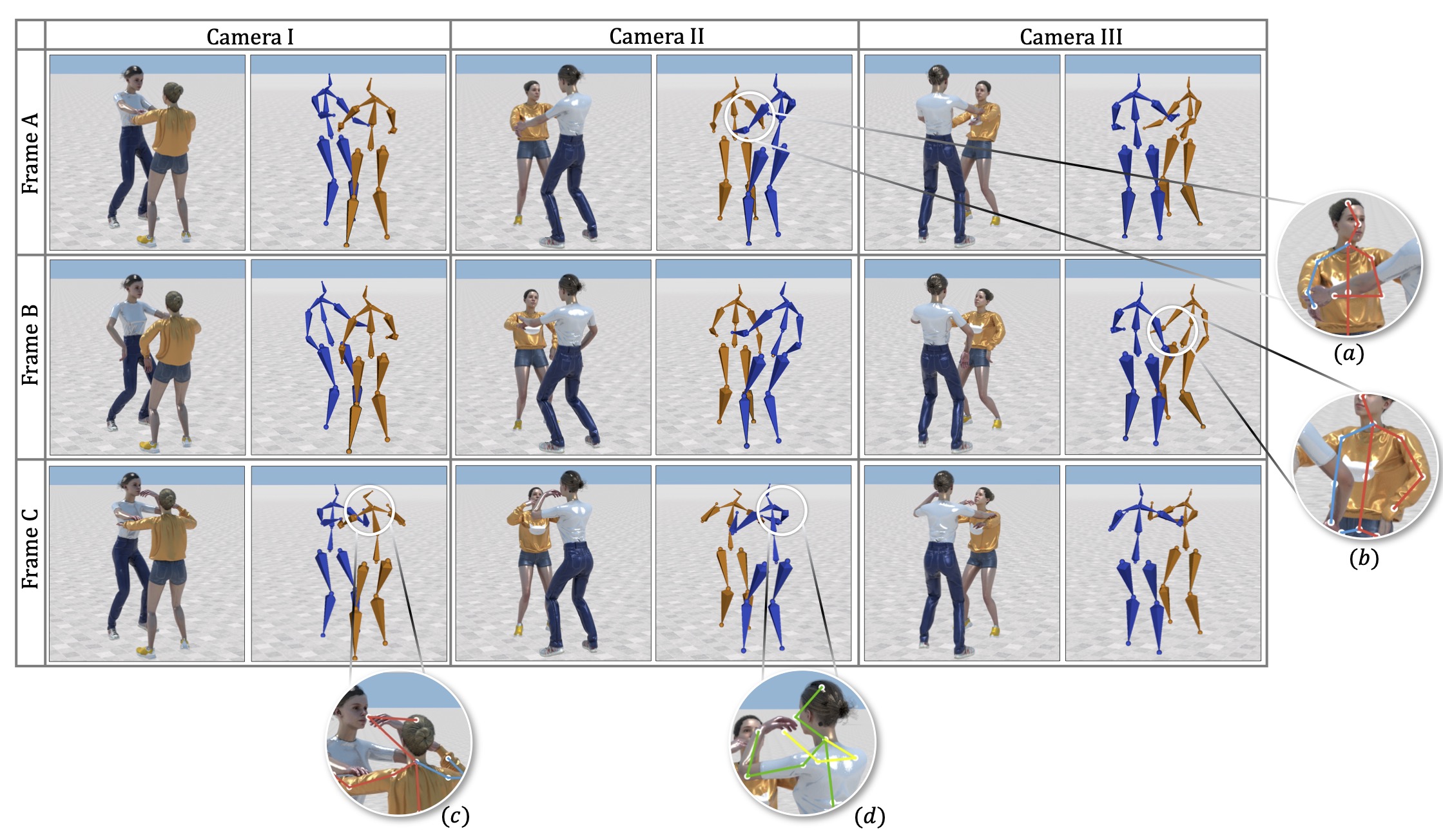}
\setlength{\abovecaptionskip}{-10pt plus 3pt minus 2pt}
\setlength{\belowcaptionskip}{-10pt plus 3pt minus 2pt}
\caption{Our results on multi-person synthetic videos, picturing two Macarena dancers. Some of the 2D joints, used as input to our method, are severely inaccurate. However, our method is able to reconstruct correct 3D motion. In the following examples, let \emph{white dancer} and \emph{orange dancer} denote the dancer wearing a white and an orange shirt respectively. Several 2D based skeleton error examples are depicted in the zoomed-in circular insets: 
(a) Wrong pose estimation of the left arm of the orange dancer;
(b) The right arm of the orange dancer is occluded hence detected erroneously; 
(c) The nose tip of the orange dancer is erroneously detected as the nose tip of the white dancer;
(d) Erroneous 2D pose estimation of the white dancer's right hand.}
\label{fig:macarena_results_with_circles}
\end{figure*}
In \Cref{fig:macarena_results_with_circles} we depict qualitative results for a scene with two macarena dancers. We emphasize several viewpoints where the 2D backbone attains large errors. Yet, FLEX is able to compensate for these errors by fusing multi-view information.
\Cref{fig:fight_2d,fig:macarena_2d} depict 2D joint locations estimated by the AlphaPose~\cite{alphapose} backbone. A close look at these figures shows that many of the estimated locations are inaccurate, e.g., a hand of one subject is confused with the hand of the other subject. Even though the number of 2D errors is large, our algorithm is able to reconstruct the characters correctly.

\sectiontinyvert{Skeleton} \label{sec:skeleton}

\begin{figure}[tbh]
\centering
\includegraphics[width=\linewidth]{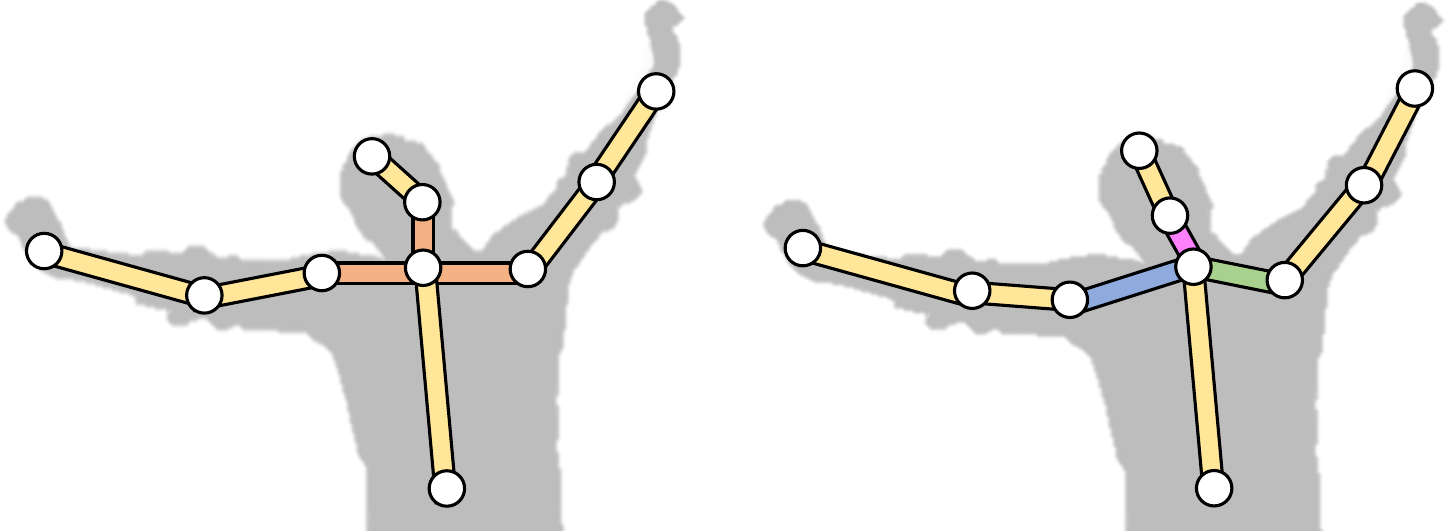}
\setlength{\abovecaptionskip}{-5pt plus 3pt minus 2pt}
\setlength{\belowcaptionskip}{-3pt plus 3pt minus 2pt}
\caption{Skeletal connectivity changes, demonstrated on the neck joint. Left: original connectivity, where shoulders and head are rigidly connected, yielding poor reconstruction. Right: new connectivity, with extra degrees of freedom. }
\label{fig:skeletal_change}
\end{figure}

To better reconstruct the motion in a given video stream, we modify the skeleton connectivity used in our baseline~\cite{shi2020motionet} (\Cref{fig:skeletal_change}).
The root and neck joints of the baseline skeleton are both rigidly attached to the three bones neighboring each of them.
This rigid connectivity constrains the skeleton, e.g., a motion where each shoulder moves forward and the neck moves to the right is impossible.
In order to remove this constraint we add joints that overlap the root and the neck, hence enabling the neighboring bones to move independently of each other. 

The new skeleton connectivity better matches the Human3.6M rotation angles ground-truth, thus, it better matches the way the dataset motions were captured.
The new skeleton improves the mean per joint position error (MPJPE) both in the multi-view setting and the monocular case. The improvements are by  $\sim$4mm and $\sim$6mm for monocular and four cameras setting, respectively.

\sectiontinyvert{Architecture details} \label{sec:architecture_details}

\begin{figure}[tbh]
\centering
\includegraphics[width=\linewidth]{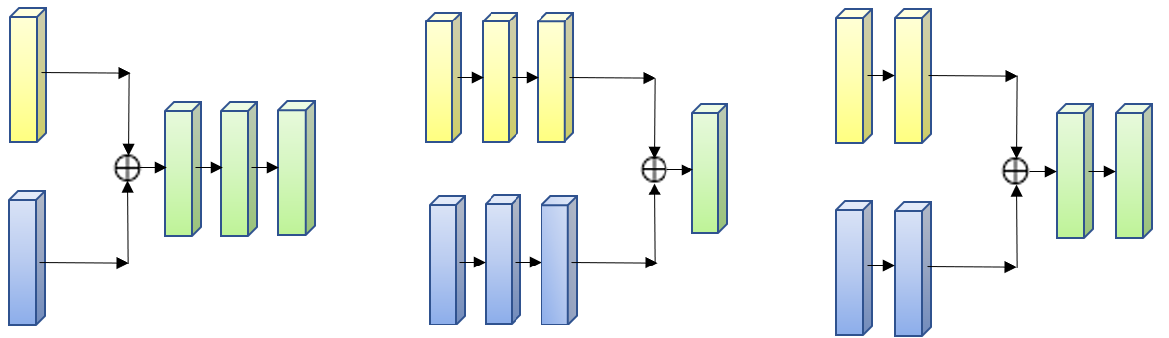}
\setlength{\abovecaptionskip}{-10pt plus 3pt minus 2pt}
\setlength{\belowcaptionskip}{-15pt plus 3pt minus 2pt}
\caption{Fusion schemes. Left: early fusion; Middle: late fusion; Right: middle fusion. Yellow and blue blocks symbolize features from distinct input branches, and green blocks represent fused data. Each block stands for a tensor of features. }
\label{fig:fusion}
\end{figure}
\begin{figure*}[!th]
\centering
\includegraphics[width=\linewidth]{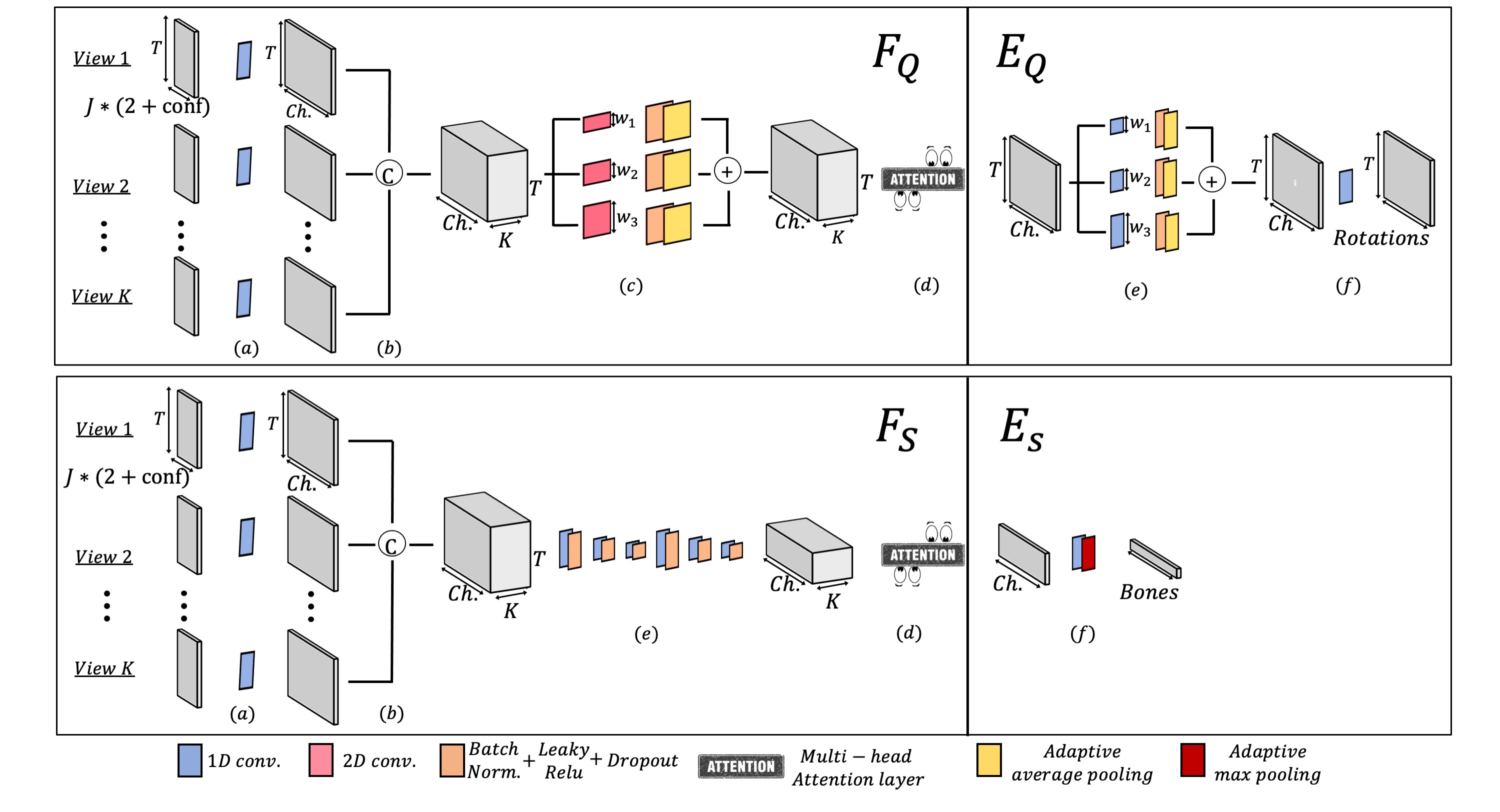}

\setlength{\abovecaptionskip}{0pt plus 3pt minus 2pt}
\setlength{\belowcaptionskip}{10pt plus 3pt minus 2pt}

\caption{Architecture in detail. The upper and lower parts are the rotations and bones branches, respectively.  (a) Channel-wise expansion layer; (b) View concatenation; (c) Multi-view convolutional filters; (d) Cross-view attention layer \bg{(detailed in \Cref{fig:attention_detail})}; (e) Single-view convolutional filters; (f) Channel-wise shrinkage layer.}
\label{fig:architecture_detail}
\end{figure*}
\begin{figure*}[!th]
\centering
\includegraphics[width=\linewidth]{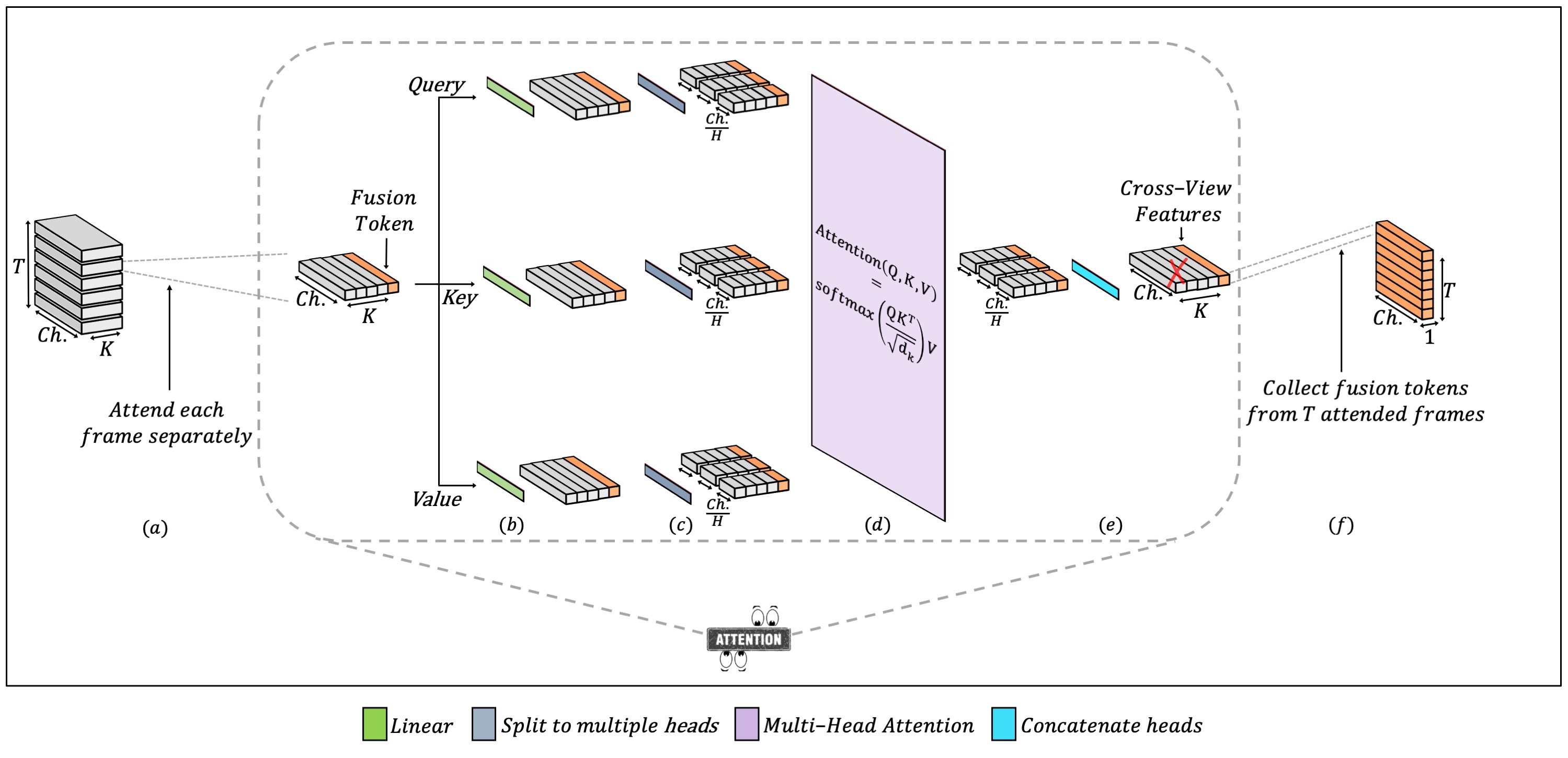}

\setlength{\abovecaptionskip}{5pt plus 3pt minus 2pt}
\setlength{\belowcaptionskip}{-10pt plus 3pt minus 2pt}
\caption{\bg{Cross-view attention layer (\Cref{fig:architecture_detail}(d)) in detail. Our attention mechanism processes each frame separately, attending the multi-view features and fusing them to a single output per frame. 
(a)~Process each temporal frame independently. Add a learned token~\cite{BERT} that forms a \emph{fusion view}; 
(b)~Linear layer; 
(c)~Split the channels to $H$ attention heads; 
(d)~Multi-Head attention~\cite{attn_is_all_you_need}; 
(e)~Concatenate the attention heads; 
(f)~Drop features from the original views. Collect fusion view features from all the frames in the temporal sequence. }}
\label{fig:attention_detail}
\end{figure*}
\begin{table*}[bht]
\setlength{\abovecaptionskip}{5pt plus 3pt minus 2pt}
\setlength{\belowcaptionskip}{0pt plus 3pt minus 2pt}
\caption{FLEX structure. $\bJ$ denotes the number of joints, $\bK$ the number of views, and $\bL$ denotes the number of limbs. $k$ and $s$ denote kernel width and the stride, respectively. $\rightarrow$ denotes parallel convolutions while $\downarrow$ denotes sequential ones.}
\resizebox{\textwidth}{!}{
\begin{tabular}{| l | c | l | c | c | c |}
\hline
\toprule
Name & & Layers & $k$ & $s$ & in\; /\; out \\
\hline
\midrule 
$F_Q$  & &1D-Conv + BatchNorm + LReLU + Dropout  & $1$ & 1 & $3J\; /\; 1024$\\
\cline{2-5}
& $\rightarrow$ &2D-Conv + BatchNorm + LReLU + Dropout + Adap AP & $5$ & 3 & $1024\; /\; 1024$\\
& $\rightarrow$ &2D-Conv + BatchNorm + LReLU + Dropout + Adap AP & $3$ & 1 & $1024\; /\; 1024$\\
& $\rightarrow$ &2D-Conv + BatchNorm + LReLU + Dropout + Adap AP & $1$ & 1 & $1024\; /\; 1024$\\
\cline{2-5}
& & Multi-head Attention layer (64 heads) & $-$ & $-$ & $1024\; /\; 1024$\\

\hline
$E_Q$ & $\rightarrow$ &1D-Conv + BatchNorm + LReLU + Dropout + Adap AP & $5$ & 3 & $1024\; /\; 1024$\\
& $\rightarrow$ &1D-Conv + BatchNorm + LReLU + Dropout + Adap AP & $3$ & 1 & $1024\; /\; 1024$\\
& $\rightarrow$ &1D-Conv + BatchNorm + LReLU + Dropout + Adap AP & $1$ & 1 & $1024\; /\; 1024$\\
\cline{2-5}
& & 1D-Conv & $1$ & 1 & $1024 / 4(J\!-\!1)\!+\!4K$\\
\hline
\midrule 
$D$ & & 1D-Conv + ReLU  & $1$ & 1 & $4J \; /\; 1024$\\
& $\downarrow$&1D-Conv + ReLU + Adap AP & $1$ & 1 & $1024\; /\; 24J$\\
& &Linear & $-$ & $-$ & $24J \; /\; J$\\
\hline
\midrule 
$F_S$ & & 1D-Conv + BatchNorm + LReLU + Dropout  & $1$ & 1 & $J3\; /\; 1024$\\
\cline{2-5}
& &1D-Conv + BatchNorm + LReLU + Dropout & $5$ & 1 & $1024\; /\; 1024$\\
& $\downarrow$ &1D-Conv + BatchNorm + LReLU + Dropout & $3$ & 1 & $1024\; /\; 1024$\\
& &1D-Conv + BatchNorm + LReLU + Dropout & $1$ & 1 & $1024\; /\; 1024$\\
\cline{2-5}
& &1D-Conv + BatchNorm + LReLU + Dropout & $5$ & 1 & $1024\; /\; 1024$\\
& $\downarrow$ &1D-Conv + BatchNorm + LReLU + Dropout & $3$ & 1 & $1024\; /\; 1024$\\
& &1D-Conv + BatchNorm + LReLU + Dropout & $1$ & 1 & $1024\; /\; 1024$\\
\cline{2-5}
& & Multi-head Attention layer (64 heads) & $-$ & $-$ & $1024\; /\; 1024$\\

\hline
$E_S$ & &Adaptive MP & $-$ & $-$ & $-$\\
& &1D-Conv & $1$ & 1 & $1024\; /\; L$\\
\hline
\bottomrule
\end{tabular}
} 
\setlength{\abovecaptionskip}{0pt plus 3pt minus 2pt}
\setlength{\belowcaptionskip}{-20pt plus 3pt minus 2pt}
\caption*{}
\label{tab:layer_detail}
\end{table*}

The architectural blocks in our implementation are the multi-view feature fusion layers $F_S$ and $F_Q$, the two encoders, $E_S$ and $E_Q$, a forward kinematics layer $FK$ and a discriminator $D$.
Our discriminator $D$ is a linear component that contains two convolution layers and one fully connected layer. We base it on Kanazawa~\etal~\shortcite{Kanazawa:2018}). 
The $FK$ layer is based on Villegas \etal~\shortcite{Villegas:2018}. 

There are two novel building blocks contained in the new fusion layers, $F_S$ and $F_Q$. The first is a multi-view convolutional layer; that is, a convolution that is aware of features stemming from  multiple views as well as multiple frames. This convolutional layer is used in $F_Q$ only.
The second is a multi-head attention layer, used in both $F_S$ and $F_Q$.
\sr{A standard attention mechanism looks at the data as a \emph{sequence} of \emph{embeddings}. In our mechanism, the \emph{views} form the sequence, and the \emph{channels} form the embeddings.}
Our attention layer is based on the LiftFormer~\cite{llopart2020liftformer}. While the LiftFormer attends to time, we attend to views. 
\sr{The embedding size is 1024, and we use 64 heads (see \Cref{tab:layer_detail}), so for the Query, Key and Value (each), we have 64 learned linear filters of size $K\times 16$, where $K$ is the number of views and 16 is the result of 1024/64.}

A key architectural choice in our fusion layers, $F_S$ and $F_Q$, is at which stage to fuse the input views. In \Cref{fig:fusion} we depict the conceptual idea of fusing. Each fusing scheme has its own advantages and disadvantages. Following the insight that early convolutional layers yield coarse features and late ones yield semantic features, we observe early fusion as generating all features (coarse and semantic) when a network is aware to all input branches, and observe middle fusion as first generating coarse features that are distinct for each branch, and only then fuse the coarse features together to generate common semantic features. When applying late fusion, the network creates distinct coarse and semantic features for each branch and only then fuses them together.
During the development of our model, we have experimented with different fusion schemes, and found out that for our setting the early fusion works best.

\Cref{fig:architecture_detail} depicts diagrams of the multi-view fusion layers and the encoders. The input to both fusion layers is $\bV_{s,q,r}\in\bbr^{T\times 3J\times K}$ (described in \ifappendix{\Cref{sec:architecture}}\else{the Architecture section of the main paper})\fi. 
In order to make the diagram more intuitive, we sketch $V$ as $K$ temporal sequences. 
Each temporal sequence is a 2D tensor, where channels are formed by the joints.
The fusion layer first streams these temporal sequences through an expansion layer, increasing their channel size. 
Next, our fusion layer concatenates the expanded data and obtains a 3D tensor, on which it applies multi-view convolutional filters. These filters consider the data from all views. At the next stage we apply a multi-head attention layer that attends to views. Our network uses the attention layer output to create a 2D tensor representing one 'fused' view. The features are then passed to the encoder. 
The encoder block $E_Q$ consists of three parallel 1D convolutional layers of different kernel sizes, followed by a final additional 1D convolution. The encoder $E_S$ starts with an adaptive max pooling to collapse the time dimension and then runs a final 1D convolution.
After each convolution block, we apply batch normalization, a leaky rectified linear unit and dropout. 
Finally, we run a shrinking filter to decrease the number of channels to the desired output size. 
Table~\ref{tab:layer_detail} describes the weight parameters of each layer.

\sr{
In \Cref{fig:attention_detail} we zoom into the attention block (item (d) in \Cref{fig:architecture_detail}). 
Each slice of one temporal frame is separately forwarded through this block. 
Such a slice contains features from all the views. Within the attention block, we concatenate an additional view, which we call the \emph{fusion view}. 
This additional view is a learned token~\cite{BERT}, in which the attention mechanism combines significant information from all views. 
Our model attends to all views, including the added one.
After exiting the attention block we omit the data related to the given views and keep the learned fusion view only.
This fusion view is then concatenated with the outputs of the other temporal frames.
}

\sectiontinyvert{Data}  \label{sec:data}
\subsectiontinyvert{Train and Evaluation} 
We train our model on the Human3.6M dataset~\cite{h36m_pami,IonescuSminchisescu11}. We evaluate FLEX on the Human3.6M and the KTH Multi-view Football II~\cite{footballDS} datasets, and on synthetic multi-person video streams captured by dynamic cameras.

Human3.6M~\cite{h36m_pami,IonescuSminchisescu11} is a dataset of 3.6 Million accurate 3D Human poses, acquired by recording the performance of 5 female and 6 male subjects, under 4 different viewpoints.
This dataset holds a diverse set of motions and poses encountered as part of 17 typical human activities such as talking on the phone, walking, and eating.
As recommended on the Human3.6M dataset page, we use subjects S1, S5, S6, S7, and S8 for training and subjects S9 and S11 for testing.
This dataset grants licenses free of charge that are limited to academic use only. More information and access to raw data are provided on the dataset webpage\footnote{\url{https://vision.imar.ro/human3.6m/}}.

KTH Multi-view Football II~\cite{footballDS} is a dataset of video streams from three synchronized cameras with 800-time frames per camera. The streams depict two different players (in separate streams), where each player has two sequences in varying levels of scene complexity.
This dataset is unique in the sense that the cameras are dynamic, hence the approximation of camera extrinsic parameters is very challenging.
We adjust the skeleton topology of the KTH dataset to match the topology of Human3.6M in the following way. KTH extracts 14 joints (top-head, mid-head, shoulders, hips, knees, feet, elbows, and hands). We create root (pelvis) and neck joints by averaging the hips and the shoulders respectively and then create a spine joint by averaging the root and the neck. Then we draw bones according to the Human3.6M skeleton topology.
The raw data can be accessed and downloaded from the dataset webpage\footnote{\url{https://www.csc.kth.se/cvap/cvg/?page=footballdataset2}}. This dataset may only be used for academic research and not for commercial use.

Ski-Pose PTZ-Camera~\cite{ski_ptz} is a 6 cameras' multi-view pant-tilt-zoom-camera (PTZ) dataset. It features competitive alpine skiers performing giant slalom runs.
It holds 20K images, with a single skier in each. 
The cameras can rotate, but their locations are fixed.
This dataset provides labels for the skiers’ 3D locations in each frame, their projected 2D locations, and per-frame calibration of the PTZ cameras.
In the dataset webpage\footnote{\url{https://www.epfl.ch/labs/cvlab/data/ski-poseptz-dataset/}} there are more descriptions of the dataset as well as download instructions.

Our synthetic videos are generated using Mixamo~\cite{mixamo} and Blender~\cite{blender}. We maintain two scenes with two interacting subjects in each. One scene illustrates a boxing arena, and one shows Macarena dancers. We create as many cameras as we wish, with full control on each camera's dynamic motion trajectory. Each synthetic camera outputs a video of the scene, taken from its view. To evaluate FLEX on these videos, we use a network that has been trained on the Human3.6M dataset, introducing satisfactory generalization abilities of our model.

\subsectiontinyvert{Input data} The input to our network is 2D joint locations per frame, accompanied by a confidence value. We train our network with several variations of input data.
\paragraphtinyvert{Ground truth 2D pose} Obviously, training with ground truth input data yields the best possible results. We use the 2D labeling provided by the Human3.6M dataset. 

\paragraphtinyvert{Estimated 2D pose} To simulate dynamic capture environments, where 2D labels are not available, we use several state-of-the-art 2D pose estimators as 2D backbones. In our ablation studies we demonstrate the dependency on a good estimator. The estimators that we use are OpenPose~\cite{Cao:2018}, CPN~\cite{chen2018cascaded}, and the one used by Iskakov \etal~\shortcite{iskakov2019learnable} (who base their 2D estimation on the ”simple baselines” architecture~\cite{xiao2018simple}). 
OpenPose and Iskakov \etal provide confidence values that we add to the network input. CPN does not provide these values, hence we assign identical confidence values for all joints when using it.
While OpenPose and CPN are dedicated 2D pose estimators, Iskakov \etal's 2D estimator is part of a 3D pose estimator. To extract the 2D pose we retrain their model using its given code and save intermediate values.
The 2D pose estimation computed by Iskakov \etal~\shortcite{iskakov2019learnable} uses camera distortion parameters. In addition to being free of extrinsic camera parameters, we are strict about not using the intrinsic ones as well (see Section 3 in the main paper); hence, we retrain Iskakov \etal~\shortcite{iskakov2019learnable} without those parameters.

Skeleton topology may vary between the aforementioned 3D datasets and 2D poses predicted by backbone algorithms. To mitigate this, we make adjustments to the predicted 2D joints. Openpose \cite{Cao:2018} extract 16 joints (root, neck, mid-head, top-head, shoulders, hips, knees, feet, elbows, and hands). These joints exist in the aforementioned datasets as well. In addition, a spine joint, which exists only in the 3D datasets, is artificially added (calculated as the 2D spatial average between the root and the neck joint).
For the CPN \cite{chen2018cascaded} 2D prediction, we simply use the values computed by Pavllo~\etal~\cite{pavllo20193d} and provided in their project page, which already possess the requested topology. Lastly, Iskakov \etal~\shortcite{iskakov2019learnable} predict the exact joints required by the aforementioned 3D datasets.

At inference time, when videos from the wild are used, we use a network that was trained using an estimated 2D pose and make sure that during inference, the exact 2D backbone that was used for training, is applied.

\subsectiontinyvert{Ground truth}
During train time we use 3D joint location ground truth per view, plus rotation ground truth for the discriminator. 
In contrast to location ground truth, rotation ground truth is required only once, no matter how many views we have.
During test time we need none of the above.



\sectiontinyvert{Camera Parameters} \label{sec:cam_param_technical}
We next formulate the notion of camera parameters.
Consider a pinhole camera model. Such a model possesses two types of parameters, extrinsic and intrinsic. \emph{Extrinsic} parameters correspond to 
\begin{itemize}
    \item A rotation matrix $R$: a matrix of size $3\times3$ characterizing the rotation from 3D real world axes into 3D camera axes.
    \item A translation vector $T$: a vector of size  $3$ representing the translational offset of the camera in the 3D scene.
\end{itemize} 
\emph{Intrinsic} parameters, stored in a $3\times3$ matrix $K$, are specific to a camera. $K$ consists of the focal length $f_x , f_y$, the camera optical center $c_x ,c_y$ and a skew coefficient $s_k$:
\begin{equation} \label{eq:intrinsics}
    K = \begin{bmatrix}
f_x & s_k & c_x\\
0 & f_y & c_y \\
0 & 0 & 1
\end{bmatrix}.
\end{equation}
We denote the mapping from 3D world coordinates into a 2D image plane by a $3\times4$ matrix $P$. $P$ is sometimes called  \emph{camera matrix} or\emph{ projection matrix}.
To calculate $P$, both camera extrinsic and intrinsic parameters are used:
\begin{equation}
\label{P matrix}
    P = K \times \begin{bmatrix}
R \kern2pt | \kern2pt T
\end{bmatrix} .
\end{equation}

\begin{figure*}[ptbh]
\centering
\setlength{\abovecaptionskip}{0pt plus 3pt minus 2pt}
\setlength{\belowcaptionskip}{-20pt plus 3pt minus 2pt}
\caption*{}
\includegraphics[width=.9\linewidth]{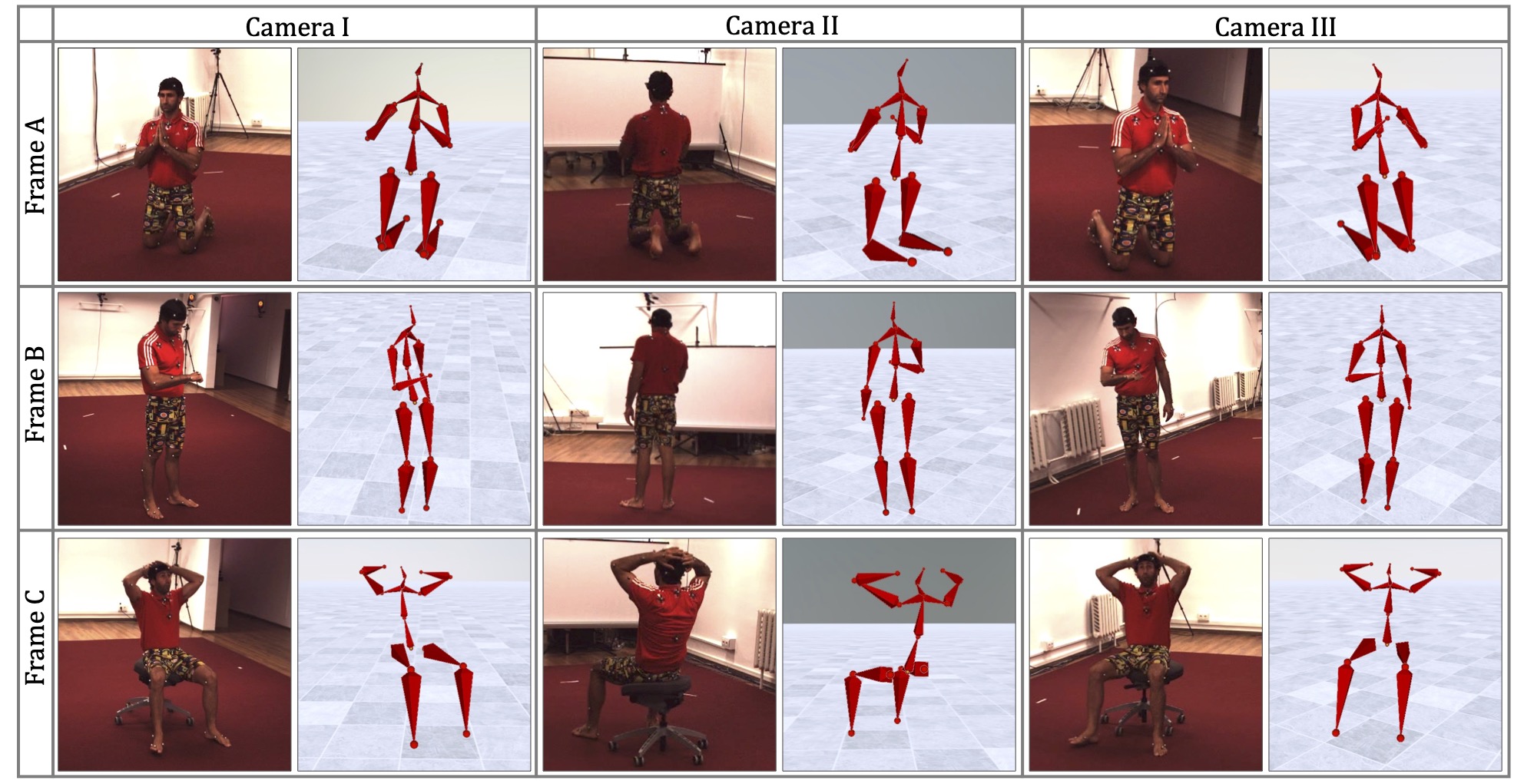}
\setlength{\abovecaptionskip}{0pt plus 3pt minus 2pt}
\setlength{\belowcaptionskip}{0pt plus 3pt minus 2pt}
\caption{Additional results on videos from the Human3.6M dataset.}
\label{fig:quality_h36_b}
\end{figure*}

\begin{figure*}[ptbh]
\centering
\includegraphics[width=.9\linewidth]{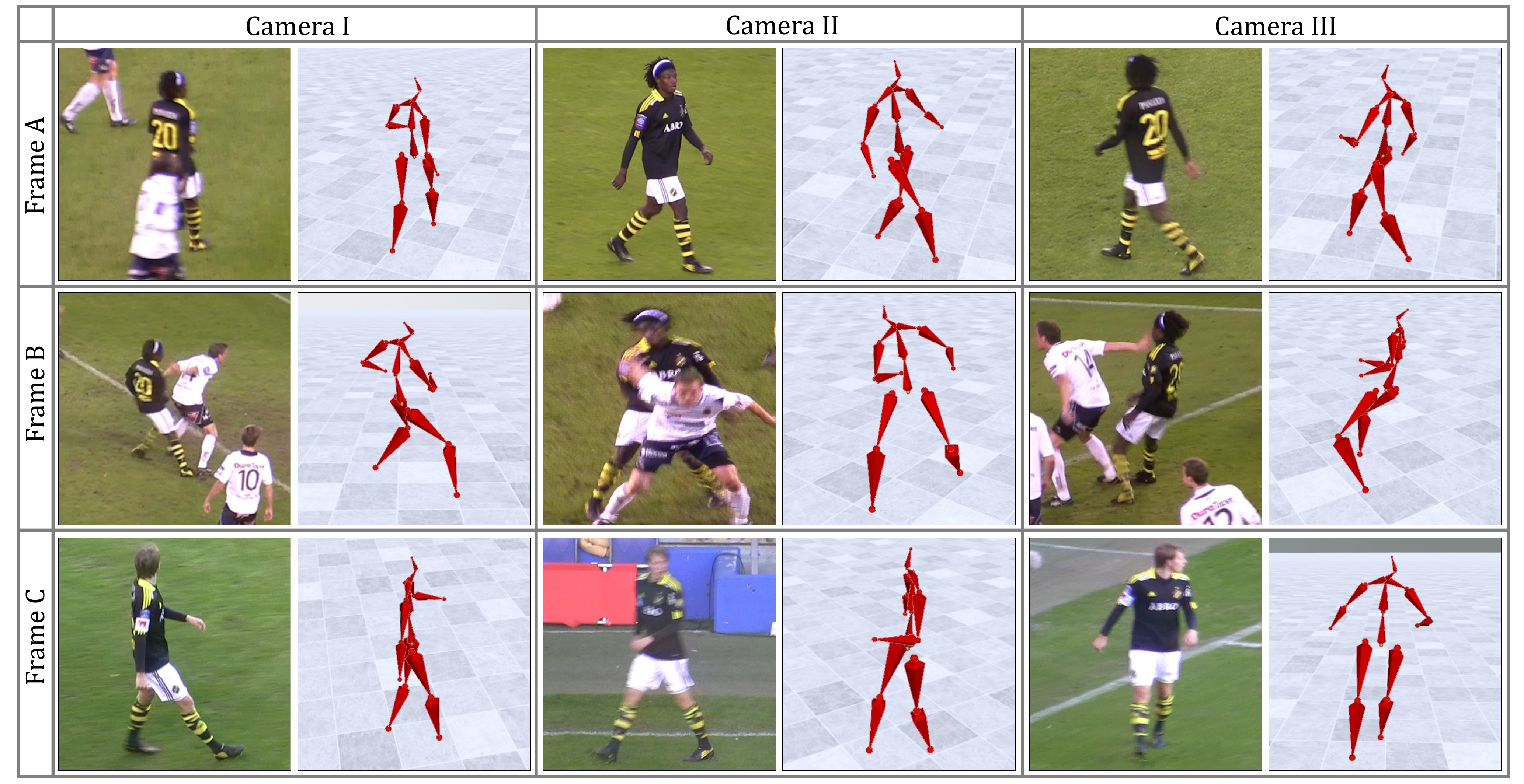}
\setlength{\abovecaptionskip}{3pt plus 3pt minus 2pt}
\setlength{\belowcaptionskip}{-7pt plus 3pt minus 2pt}
\caption{Additional results on videos from the KTH Multi-view Football II dataset. }
\label{fig:quality_KTH_b}
\end{figure*}

\begin{figure*}[tbh]
\centering
\includegraphics[width=.9\linewidth]{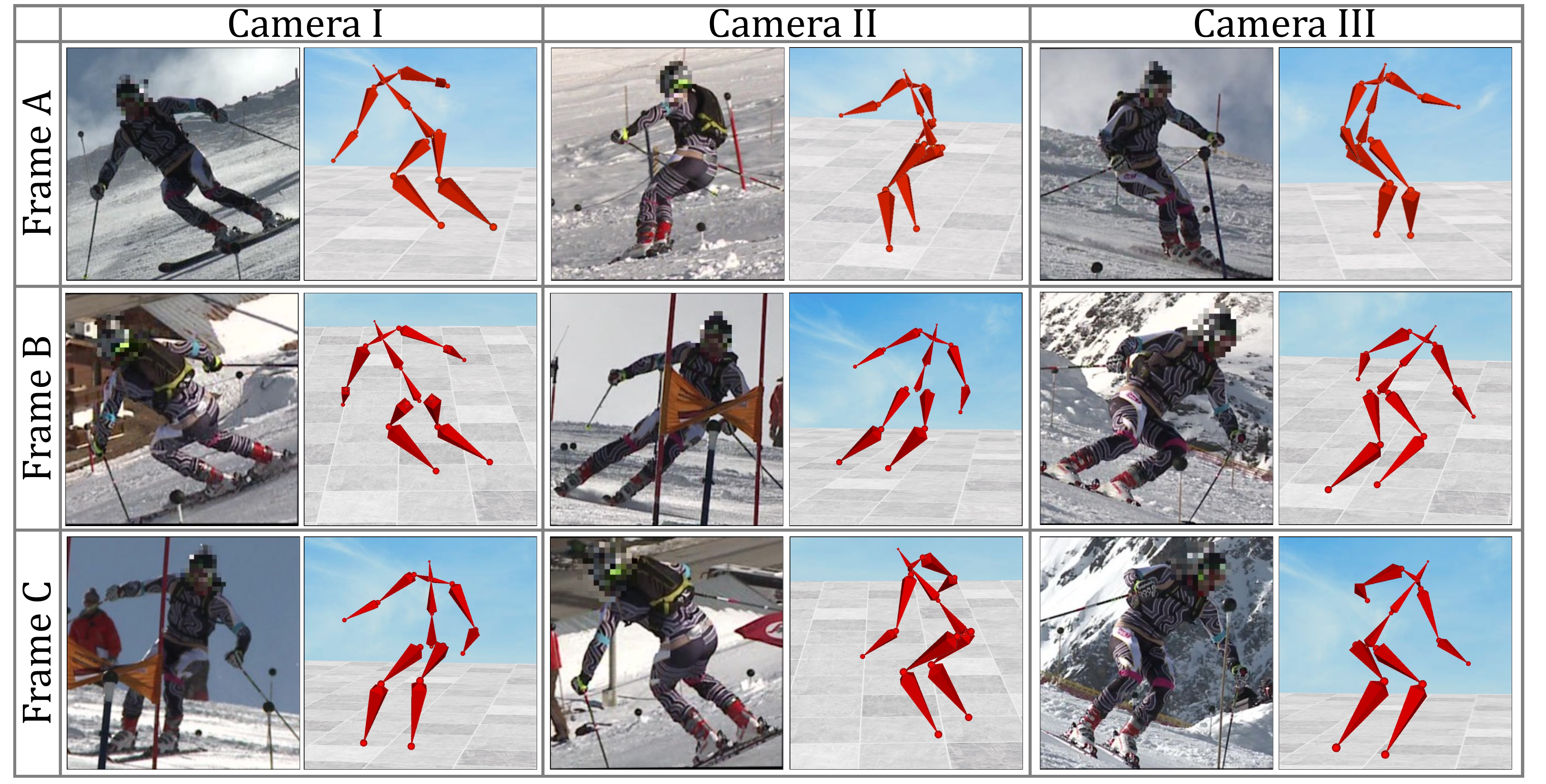}
\setlength{\abovecaptionskip}{3pt plus 3pt minus 2pt}
\setlength{\belowcaptionskip}{-7pt plus 3pt minus 2pt}
\caption{Enlarged results on the Ski-Pose PTZ-Camera dataset (from main paper).  }
\label{fig:quality_ski_large}
\end{figure*}

\begin{figure*}[tbh]
\centering
\includegraphics[width=.9\linewidth]{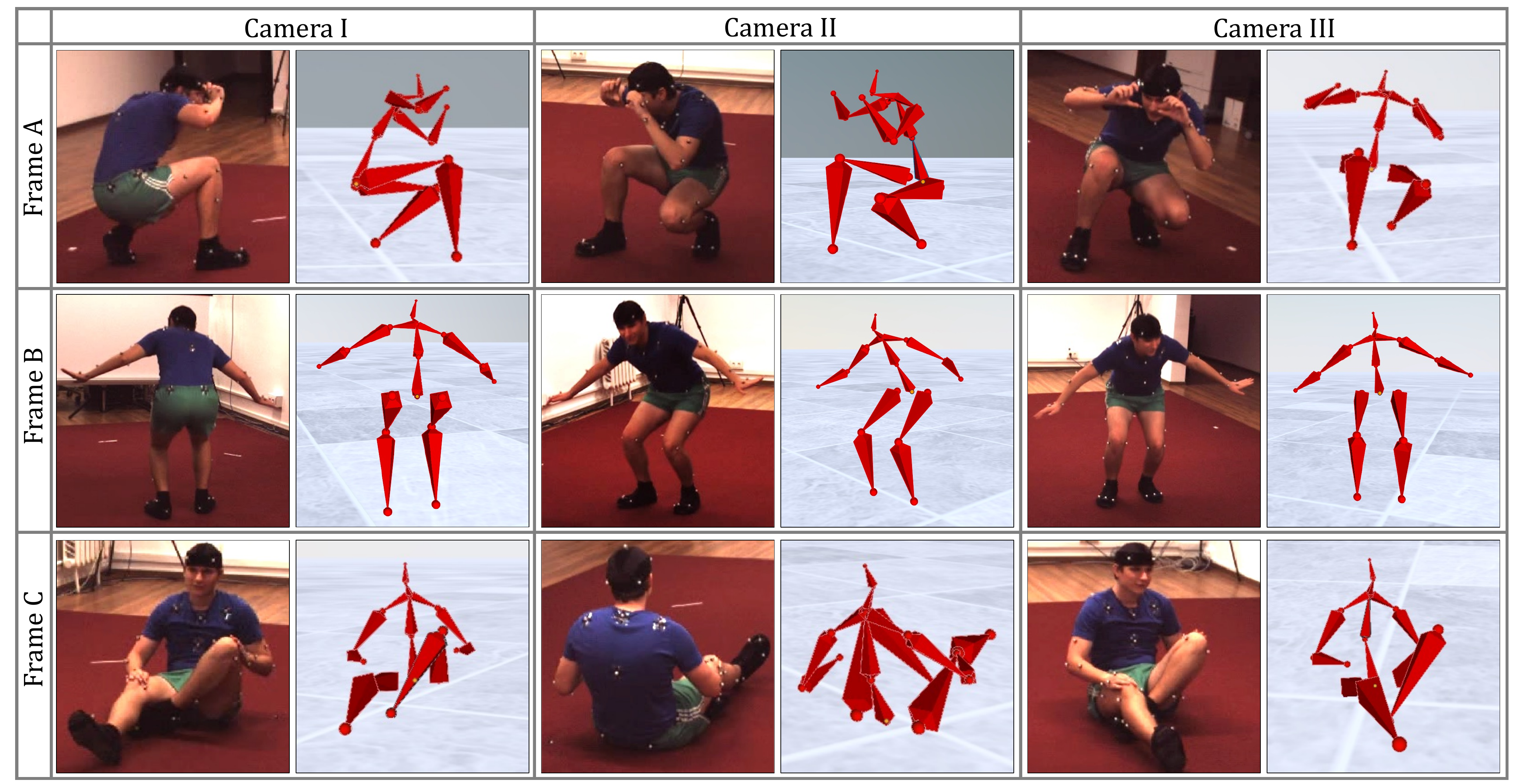}
\setlength{\abovecaptionskip}{3pt plus 3pt minus 2pt}
\setlength{\belowcaptionskip}{-7pt plus 3pt minus 2pt}
\caption{Enlarged results on the Human3.6M dataset (from main paper).  }
\label{fig:quality_h36_large}
\end{figure*}

\begin{figure*}[tbh]
\centering
\includegraphics[width=.9\linewidth]{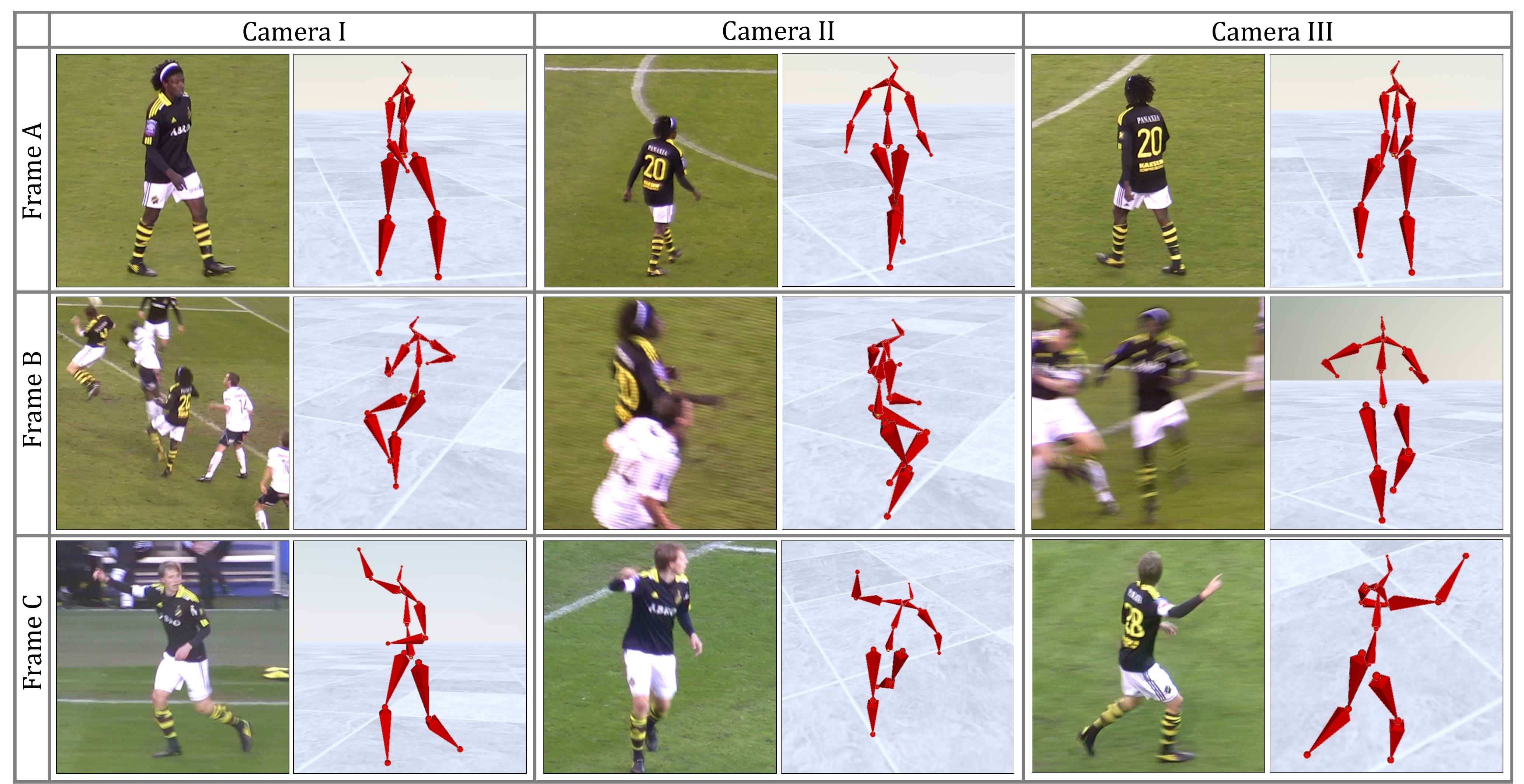}
\setlength{\abovecaptionskip}{3pt plus 3pt minus 2pt}
\setlength{\belowcaptionskip}{-7pt plus 3pt minus 2pt}
\caption{Enlarged results on the KTH Football II dataset (from main paper).}
\label{fig:quality_KTH_large}
\end{figure*}

\begin{figure*}[phtb]
\centering
\includegraphics[width=.9\linewidth]{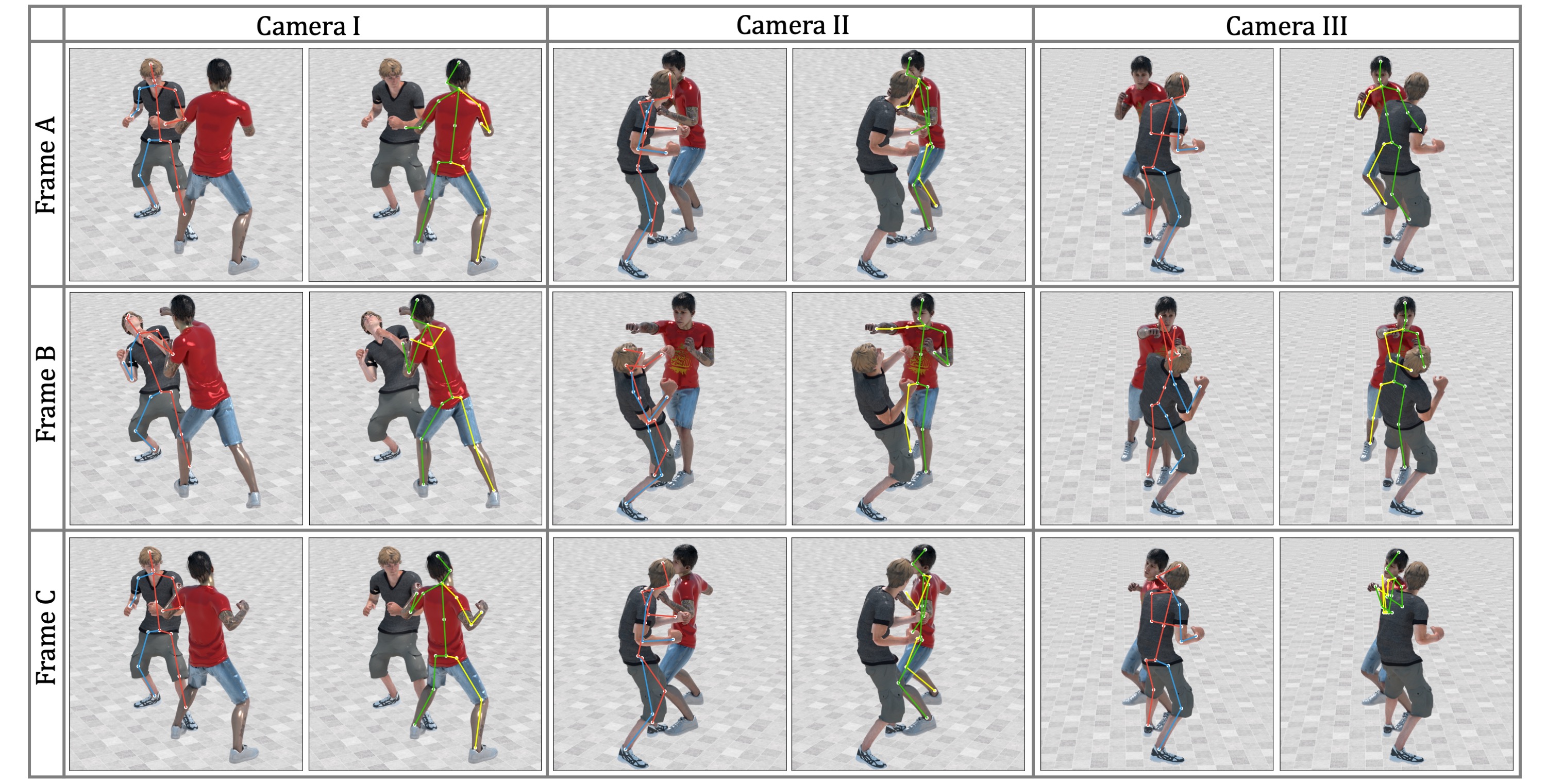}
\setlength{\abovecaptionskip}{3pt plus 3pt minus 2pt}
\setlength{\belowcaptionskip}{-7pt plus 3pt minus 2pt}
\caption{2D joint locations estimated on a multi-person synthetic video of boxers.}
\label{fig:fight_2d}
\end{figure*}
\begin{figure*}[phtb]
\centering
\includegraphics[width=.9\linewidth]{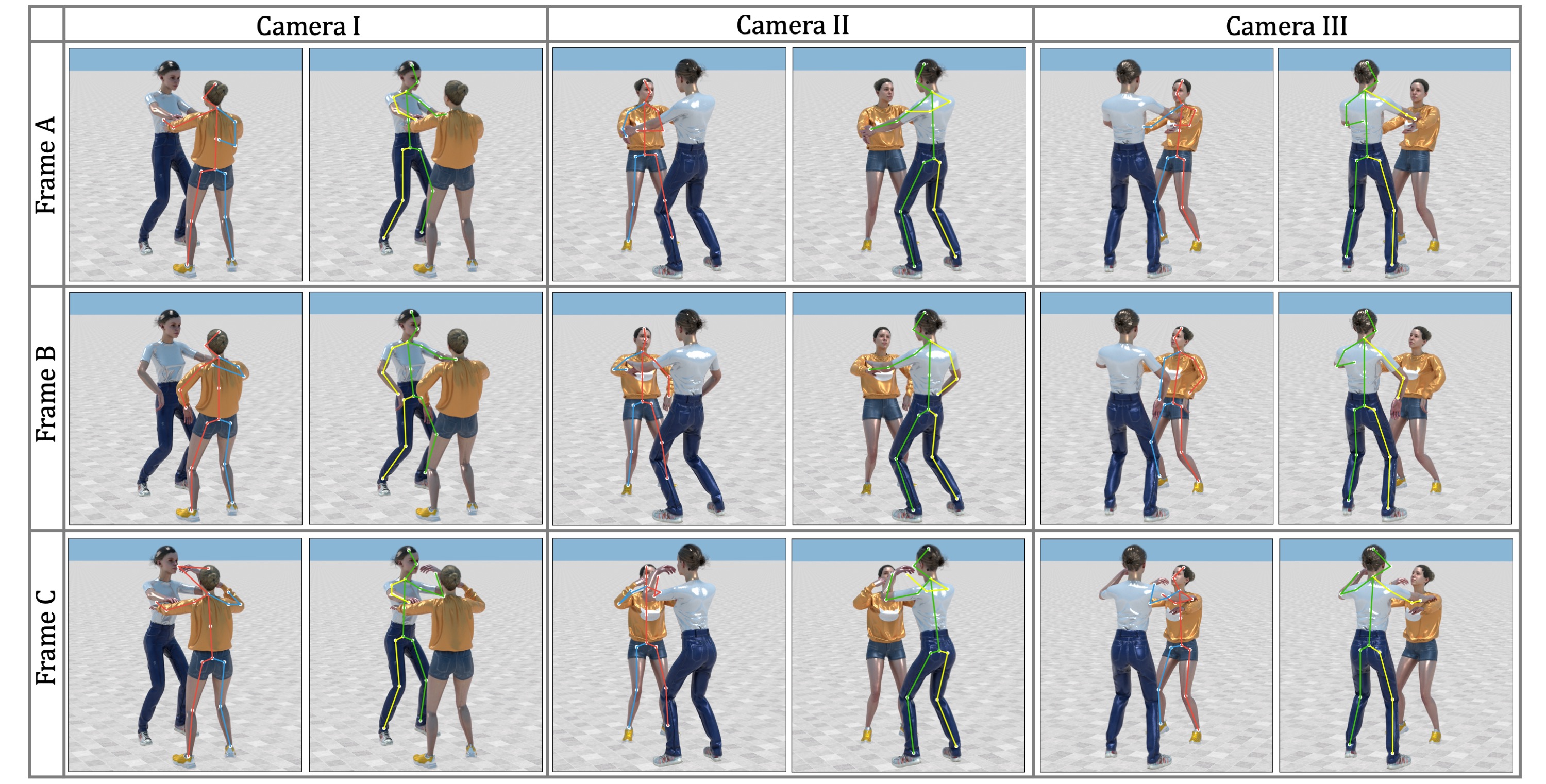}
\setlength{\abovecaptionskip}{3pt plus 3pt minus 2pt}
\setlength{\belowcaptionskip}{-7pt plus 3pt minus 2pt}
\caption{2D joint locations estimated on a multi-person synthetic video of dancers.}
\label{fig:macarena_2d}
\end{figure*}

}\fi
\end{document}